\documentclass[final,5p,times,numbers]{elsarticle}

\usepackage{amsmath, amssymb, bm}
\usepackage{xcolor}
\usepackage{soul}
\usepackage[ruled,vlined,linesnumbered]{algorithm2e}
\usepackage{placeins}
\usepackage{float}
\usepackage{booktabs}
\usepackage{multirow}
\usepackage{threeparttable}
\usepackage{adjustbox}

\usepackage[table]{xcolor}
\definecolor{lightgray}{gray}{0.85}
\usepackage{graphicx}
\usepackage{hyperref}

\SetKwInput{KwData}{Data}
\SetKwInput{KwParam}{Hyperparams}
\SetKw{KwTo}{to}
\SetKw{KwRet}{return}

\journal{Computer Methods and Programs in Biomedicine}
\usepackage{listings}

\usepackage{makecell}
\usepackage{pifont}

\begin{document}
\let\WriteBookmarks\relax
\renewcommand{\floatpagefraction}{0.7}    
\renewcommand{\textfraction}{0.15}         
\renewcommand{\topfraction}{0.85}          
\renewcommand{\bottomfraction}{0.70}       

\setcounter{topnumber}{2}
\setcounter{bottomnumber}{2}
\setcounter{totalnumber}{4}

\renewcommand{\dbltopfraction}{0.85}       
\renewcommand{\dblfloatpagefraction}{0.7}  
\setcounter{dbltopnumber}{2}

\newcommand{\SubItemStart}[1]{%
    \let\item\SubItemEnd
    \begin{SubItemList}[resume]%
        \OldItem #1%
}

\begin{frontmatter}



\title{ProteoKnight: Convolution-based Phage Virion Protein Classification and Uncertainty Analysis}
\author[inst1]{Md. Ishrak Khan\corref{cor1}\fnref{eq}}
\ead{md.ishrak.khan@g.bracu.ac.bd}
\cortext[cor1]{Corresponding author.}
\fntext[eq]{These authors contributed equally to this work.}

\author[inst1]{Abir Ahammed Bhuiyan\fnref{eq}}
\ead{abir.ahammed.bhuiyan@g.bracu.ac.bd}

\author[inst1]{Samiha Afaf Neha\fnref{eq}}
\ead{samiha.afaf.neha@g.bracu.ac.bd}

\author[inst1]{Md. Khalilur Rahman}
\ead{khalilur@bracu.ac.bd}

\author[inst2]{Jannatun Noor}
\ead{jannatun@cse.uiu.ac.bd}

\affiliation[inst1]{organization={Department of Computer Science and Engineering, BRAC University},
            addressline={Kha 224 Pragati Sarani, Merul Badda},
            city={Dhaka},
            postcode={1212},
            country={Bangladesh}}

\affiliation[inst2]{organization={C2SG Research Group, Department of Computer Science and Engineering, United International University},
    addressline={United City, Madani Ave}, 
    city={Dhaka},
    postcode={1212}, 
    country={Bangladesh}}




\begin{abstract}
\textbf{Background and Objective:} Accurate prediction of Phage Virion Proteins (PVP) is essential for genomic studies due to their crucial role as structural elements in bacteriophages. Machine learning methods have emerged for annotating phage protein sequences obtained via high-throughput sequencing. However, effective annotation requires informative sequence encoding. Our paper introduces ProteoKnight, a new image-based encoding method that addresses spatial constraints inherent in existing techniques, yielding competitive performance in PVP classification using pre-trained convolutional neural networks. Additionally, our study bridges a gap in uncertainty analysis for protein sequence classification by evaluating prediction uncertainty in binary PVP classification through the Monte Carlo Dropout (MCD) technique.\\
\textbf{Methods:} Our encoding method, ProteoKnight, adapts the classical DNA-Walk algorithm by incorporating pixel colors and adjusting walk distances to capture intricate protein features. Encoded sequences were classified using multiple pre-trained CNNs, with standard evaluation metrics. Additionally, variance and entropy measures were used to assess prediction uncertainty across proteins of various classes and lengths, forming the foundation of our investigation into protein classification and uncertainty quantification.\\
\textbf{Results:} Our experiments highlight the efficacy of our approach in binary classification, achieving prediction performance (approximately 90\% accuracy), comparable to state-of-the-art methods. Nevertheless, multi-class classification accuracy remains suboptimal. Furthermore, our uncertainty analysis unveils variability in prediction confidence influenced by protein class and sequence length, contributing novel insights to protein classification research.\\
\textbf{Conclusions:} Our study surpasses the sole existing PVP image encoding method, frequency chaos game representation (FCGR), by introducing a novel image encoding that mitigates FCGR's spatial information loss limitations. Combined with parameter-efficient CNNs, ProteoKnight enables accurate PVP prediction. Moreover, our uncertainty investigation identifies data points associated with low-confidence predictions, enhancing the comprehensiveness of our analysis.
\end{abstract}




\end{frontmatter}



\section{Introduction}
Proteins are essential macromolecules responsible for the structural and functional mechanisms of living things. Phage structural proteins are a subclass of the protein family corresponding to the bacteriophages, which are the most prevalent kind of biological creatures \cite{cobian2016viruses}. The PVPs are mainly associated with building structural constituents of the bacteriophage \cite{kabir2022large}, such as the baseplate and head as shown in Fig. \ref{fig:phage}, enabling efficient host-phage binding for genome transfer. The amino acid sequences, particularly those involved in structure synthesis, exhibit significant diversity among phages and their respective groups \cite{seguritan2012artificial}. Consequently, characterizing these sequences becomes a challenging yet crucial task, as the shortage of annotations for phage proteins has emerged as a hindrance in numerous research studies focused on phage genomics \cite{henn2010analysis}.

\begin{figure}[htbp]
    \centering
    \includegraphics[scale=0.6]{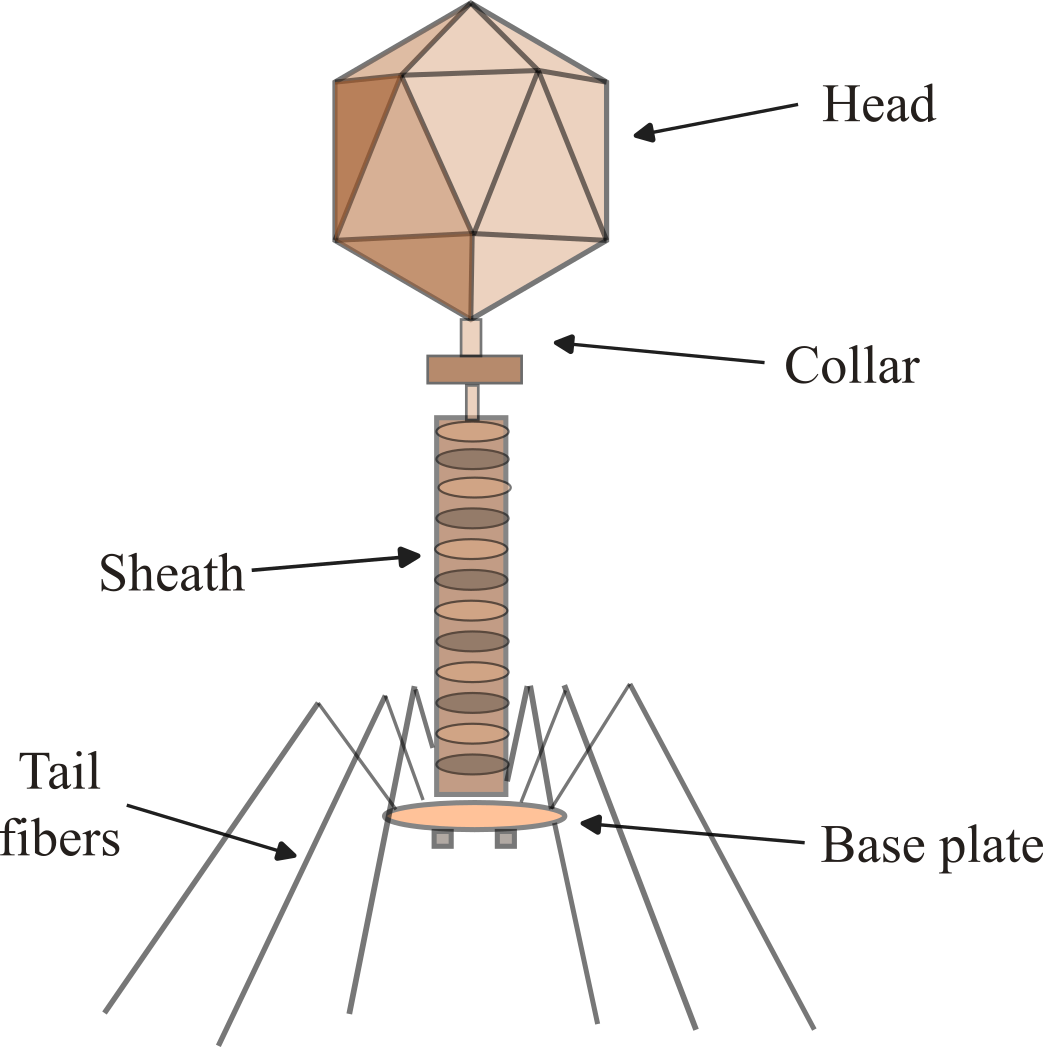}
    \caption{Structural constituents of Phage}
    \label{fig:phage}
\end{figure}

With the advent of high-throughput sequencing technology, rapid additions of sequences are being observed in standard biological databases as shown in Fig. \ref{fig:graph}, giving rise to the need for proper annotation of the sequences. Enhancement of annotations for the phage family is vital for exploring effective anti-bacterial drug synthesis \cite{meng2020review, mccallin2019current}, disease diagnosis \cite{wang2004epitope}, food production \cite{brussow2001comparative}, bacterial genome remodeling \cite{menouni2015bacterial}, etc. Traditionally, mass spectrometry and protein array techniques \cite{lavigne2009phage, yuan2016proteomic} were utilized for characterizing these proteins. However, these methods are associated with significant time, labor, and computational expenses \cite{cobian2016viruses}.

Similarly, alignment-based methods do not always work well with PVP categorization due to the lack of collinearity in viral genomes, resulting from factors such as horizontal gene transfer, high mutation rates, etc., leading to dissimilar sequences having similar structural conformations and vice versa. Thus, alignment-free methods have proved more plausible for approaching the classification task.

In light of these issues, in-silico or computational approaches leveraging machine learning techniques for faster and lower-cost classification are gaining acclaim. To enhance machine readability, amino acid sequences are initially encoded using various schemes. These encodings vary in dimension and feature selection, extracting specific features from the protein sequence necessary for identification. Subsequently, prediction models use these encoded sequences to perform their calculations.
Most traditional machine learning methods approach the PVP task as binary classification, categorizing proteins as either PVP or non-PVP \cite{arif2020pred, zhang2023rf_phage}, structural or non-structural \cite{feng2013naive}, capsid or non-capsid \cite{galiez2016viralpro}, etc. However, there remains room for improvement in classification results, particularly for protein sequences with low sequence conservation. Consequently, deep learning methods have gained popularity, demonstrating superior performance over traditional machine learning approaches across various protein classification tasks. These deep learning models not only showcase better accuracy but also incorporate more PVP classes for the classification task.

\subsection{Motivation}

Evidently, there is a constant need for developing better and more efficient classification techniques for newly sequenced phage virion proteins. An essential aspect of this task involves refining the sequence representation schemes before training the models to extract valuable local and global information. Previously, feature extraction for phages has primarily focused on one-dimensional data. While these methods yield satisfactory outcomes, the potential for improved prediction through increasing dimensionality remains largely unexplored.
Studies analyzing sequence data have demonstrated that when represented as 2-D images, the data exhibits enhanced characteristics due to increased feature space \mbox{\cite{lee2023score}}. Augmenting dimensionality typically results in improved performance and simplifies the process of feature extraction by applying a standard text-to-image encoding on sequences rather than relying on the collection of several distinct features based on the sequences' physicochemical properties, such as k-mer frequencies. Consequently, numerous studies mention utilizing image encoding techniques for biological sequences \cite{jung2023deepstabp, zhao2023deeptp}.

\begin{figure}
    \centering
    \includegraphics[width=\linewidth]{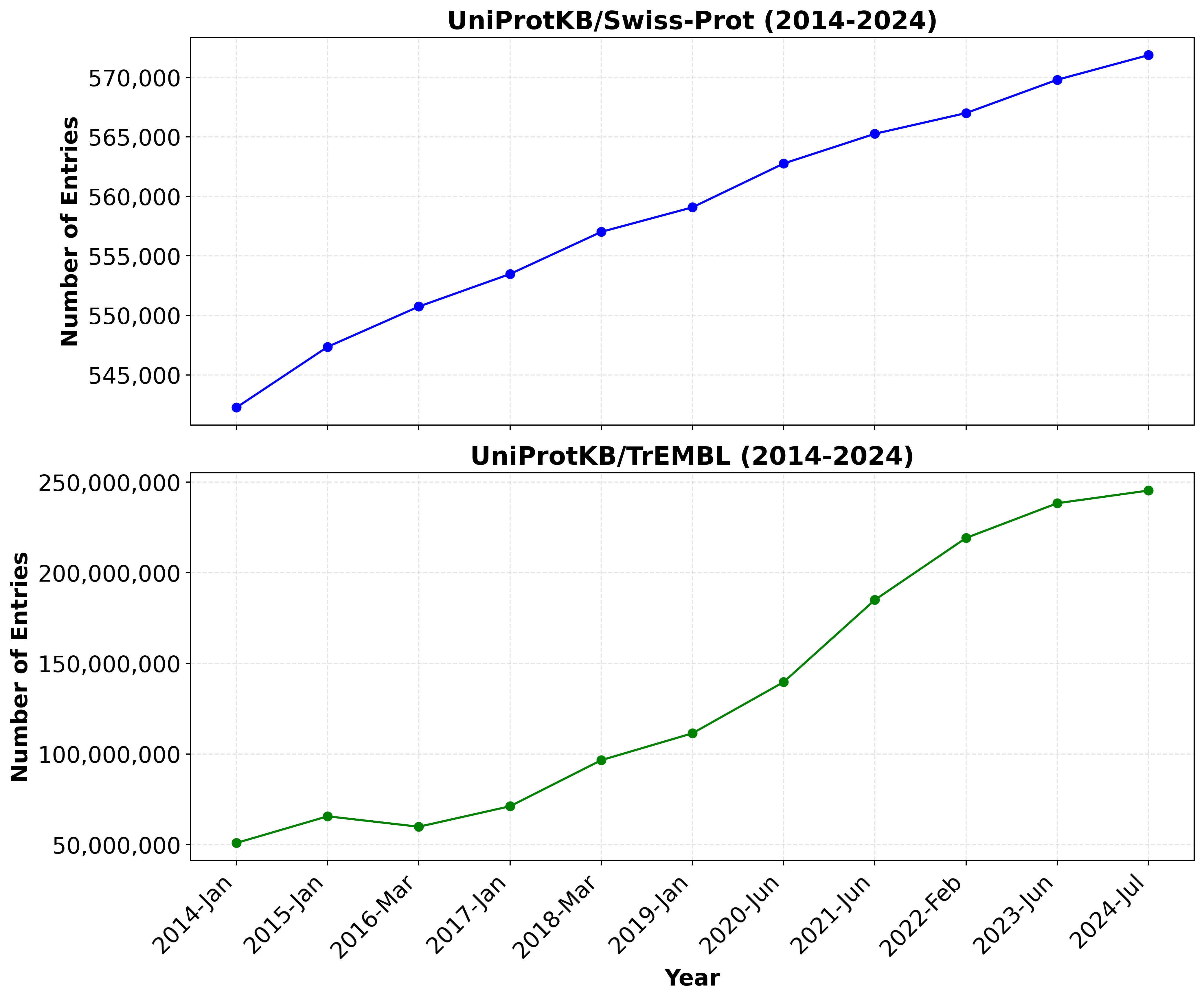}
    \caption{Growth of protein databases in past 10 years \cite{uniprot}}
    \label{fig:graph}
\end{figure}

The only study that applies image encoding for PVP classification is the one conducted in \cite{phaVIP}, which employs FCGR. However, these encoding methods based on k-mer frequency tend to remove the spatial information of the sequences because of their compact transformation \cite{akbari2022walkim}. This highlights the need to devise an alternative effective encoding strategy. Given the constraints of both earlier and contemporary encodings, there is a drive to develop innovative and more effective attributes tailored to specific objectives.

Apart from the concern of encoding, a lack of study has been observed on the impact of uncertainty in phage virion classification. Studies show that deep learning methods often give overconfident results \cite{srivastava2014dropout}, which might lead to high uncertainty in situations previously unseen by the model. Especially for PVP, where the sequence conservation is low and training data goes through a specific set of curations, there is a reasonable possibility of model uncertainty. On a contrasting note, increased accuracy of prediction is at times positively correlated with the model's uncertainty \cite{ghoshal2020estimating}. These uncertainties are particularly worrisome in safety-essential scenarios, such as medical diagnosis and drug engineering, where result reliability is of utmost importance \cite{hullermeier2021aleatoric}.
Therefore, in this study, we design a novel image-based encoding technique inspired by the DNA-walk algorithm used for DNA sequences and adapt it for phage protein sequences. This encoding has the potential to overcome the drawback of spatial encoding loss \cite{akbari2022walkim}, employing pre-trained CNNs to attain competitive results on the prediction task. Furthermore, we investigate the trend of prediction uncertainty for these protein sequences and quantify the prediction confidence in terms of sequence-varying features, providing a validation standard as to which kinds of sequences tend to give better accuracy but with lower confidence, necessitating further assessment before annotation.

\subsection{Research Contribution and Organization}
Given the existing limitations on PVP research, our study proposes ProteoKnight, a method for PVP classification via a new encoding technique, named Knight Encoding. The core contributions from our research are as follows:

\begin{itemize}
\item We proposed a novel encoding strategy, ``Knight Encoding'', to convert text-based protein sequences into image-based data.
\item We employed state-of-the-art deep learning techniques to conduct comprehensive classification analyses, evaluating the efficacy of the proposed encoding algorithm.
\item We explored uncertainty aspects in PVP classification using Monte Carlo Dropout (MCD), identifying sequence types exhibiting high accuracy yet comparatively low confidence, necessitating further assessment before annotation.
\end{itemize}

The advancements in PVP classification through ProteoKnight hold the potential for far-reaching implications across multiple scientific domains. By offering accurate and efficient classification, our model serves as a versatile tool that bridges gaps in various research and application areas. From accelerating phage therapy development in medicine, host prediction, and taxonomy classification to enhancing metagenomic analyses in environmental microbiology, ProteoKnight's utility spans a broad spectrum of biological sciences. Importantly, ProteoKnight's novel encoding approach, when used in conjunction with existing methods, offers a complementary view of sequence classification. This synergy between different approaches has the potential to capture nuanced features of protein sequences, ultimately leading to more robust and comprehensive classification results. By exploring a new encoding method for PVP classification and addressing prediction uncertainty, ProteoKnight contributes to both computational biology and practical applications, offering a fresh perspective that could be valuable in safety-critical areas like phage therapy and bio-engineering.

To fully explore these contributions and their implications, the remainder of this paper is structured as follows: Section 2 provides the literature review, respectively, contextualizing our work within the field of PVP classification. Section 3 details our methods, including the novel Knight Encoding strategy and ProteoKnight implementation. Section 4 presents our experimental results and performance evaluation, while Section 5 offers a comprehensive discussion of our findings and their implications. Finally, Section 6 concludes the paper, summarizing key contributions and outlining future research directions.

\subsection{Pre-trained CNNs for Classification}

Pre-trained convolutional neural networks (CNNs) have gained significant attention in the field of bioinformatics, particularly for protein sequence classification tasks. These models, which are initially trained on large-scale datasets such as ImageNet \cite{russakovsky2015imagenet}, have demonstrated remarkable performance in various computer vision applications. The success of pre-trained CNNs can be attributed to their ability to learn rich, transferable feature representations that can be fine-tuned for specific tasks, including phage virion protein (PVP) classification.

A major advantage of pre-trained CNNs is their computational efficiency. Training deep learning models from scratch can be time-consuming and resource-intensive, especially when dealing with large datasets. By using pre-trained models, it is possible to significantly reduce the training time and computational resources required, as the models have already learned meaningful features from extensive training on diverse datasets \cite{kornblith2019better}. This efficiency allows researchers to focus on fine-tuning the models for specific tasks and iterating on different architectures and hyperparameters.

Furthermore, the strong generalization capabilities of pre-trained CNNs make them a good choice for phage classification tasks, given their high diversity and low sequence conservation. Pre-trained CNNs have demonstrated their ability to generalize well to unseen data, making them suitable for handling the challenges associated with phage protein classification \cite{zhuang2020comprehensive}.

\subsection{Uncertainty Estimation in Prediction}

Although existing deep learning models have been seen to provide impressive accuracy, no study has yet been conducted on the impact of uncertainty in phage virion classification. Studies show that deep learning methods often give overconfident results \cite{srivastava2014dropout}, which might lead to high uncertainty in situations previously unseen by the model. On a contrasting note, increased accuracy of prediction is at times positively correlated to the model's uncertainty \cite{ghoshal2020estimating}. The aforementioned uncertainty can be categorized into two distinct types: Aleatoric Uncertainty (data-centric), and Epistemic Uncertainty (model-centric). The nature of the two uncertainties is illustrated in Figure \ref{fig:aleatoric_epistemic_uncertainty}.

\begin{figure}[htbp]
    \centering
    \includegraphics[width=\columnwidth]{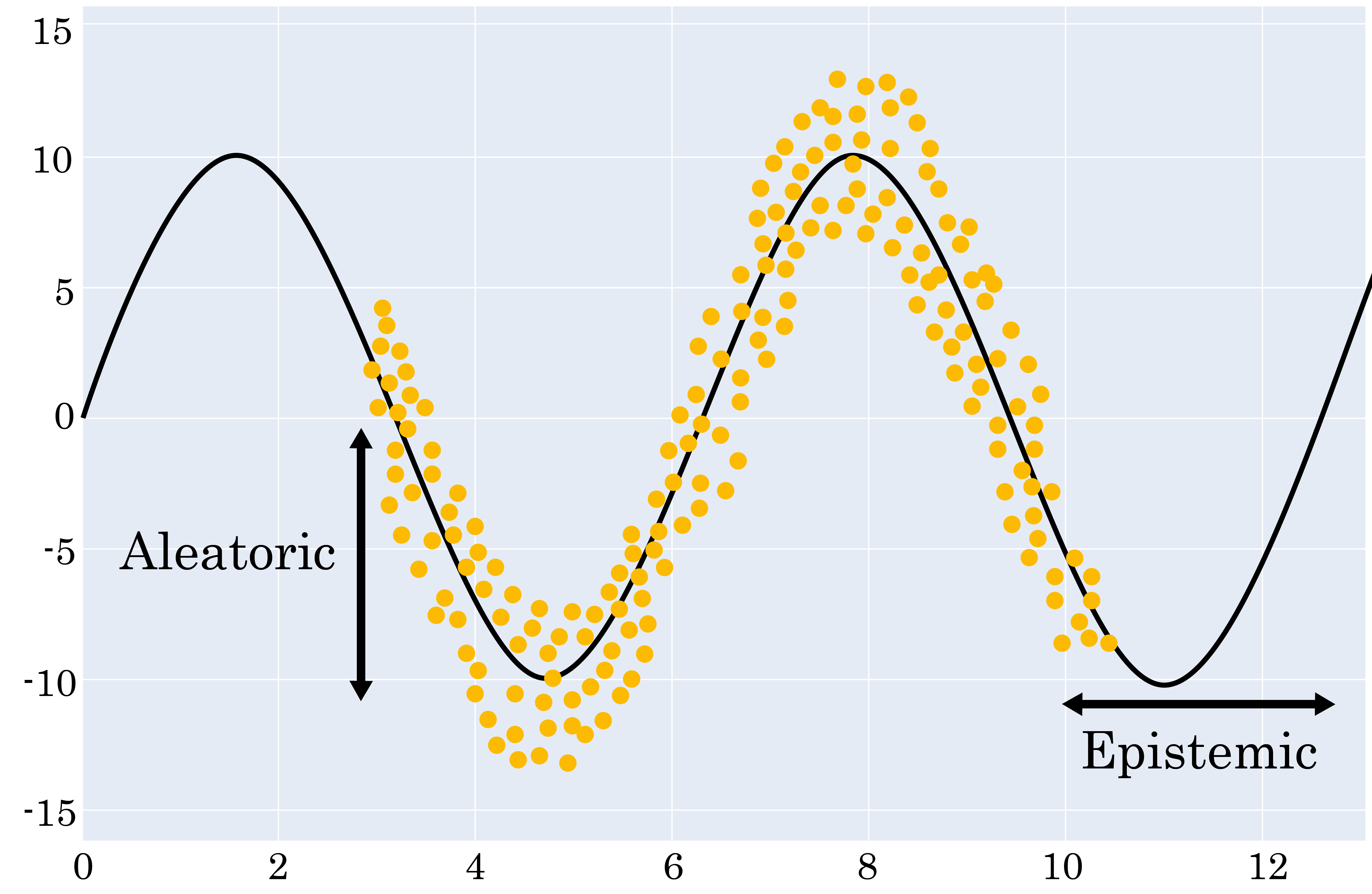}
    \caption{A schematic view of aleatoric and epistemic uncertainty in prediction \cite{abdar2021review}.}
    \label{fig:aleatoric_epistemic_uncertainty}
\end{figure}

These uncertainties are particularly worrisome in safety-critical scenarios, such as medical diagnosis and drug engineering, where result reliability is of utmost importance \cite{hullermeier2021aleatoric}. Rather than having point estimates for a task, uncertainty analysis attempts to deduce the variance over the distribution of the data in consideration. This analysis provides insight into both the reliability and accuracy of the model. Although much work has been done in uncertainty analysis for deep learning frameworks, the tasks dealt with are primarily based on either computer vision, image processing, or natural language processing targeting human language. In cases of protein sequences depicting biological language, studies regarding uncertainty have been a field yet to be explored. Especially for PVP, where the sequence conservation is low and training data goes through a specific set of curations, there is a reasonable possibility of model uncertainty. Exploring this uncertainty can establish a validation standard to detect sequence types that offer higher accuracy but lower confidence, necessitating additional evaluation before annotation.

\subsubsection{Monte Carlo Dropout}

The concept of utilizing dropout was introduced by Gal and Ghahramani \cite{gal2016dropout}, who employed it as a means of approximating probabilistic Bayesian models for deep Gaussian processes. Dropout is a regularization technique that mitigates overfitting. Bayesian neural networks aim to acquire knowledge about the posterior distribution of weights, conditioned on a given input sample. However, computing these posteriors analytically is infeasible, so sampling techniques can be used to estimate the weight posteriors. This involves conducting several stochastic evaluations utilizing distinct weight samples within the model.

Monte Carlo Dropout is such a sampling technique used to estimate the weight posterior in a given dataset. During training, it randomly retains or drops nodes inside the neural network based on a Bernoulli random variable's probability distribution \emph{P} (refer Equation \ref{ber}). As shown in Figure \ref{fig:dropout}, the dropout approach incorporates this probability parameter. Consequently, when the same input passes through the model, the output may vary slightly depending on which nodes are activated or dropped. Each of these outputs represents a distinct sample within the network.

\begin{figure}[htbp]
    \centering
    \includegraphics[width=\columnwidth]{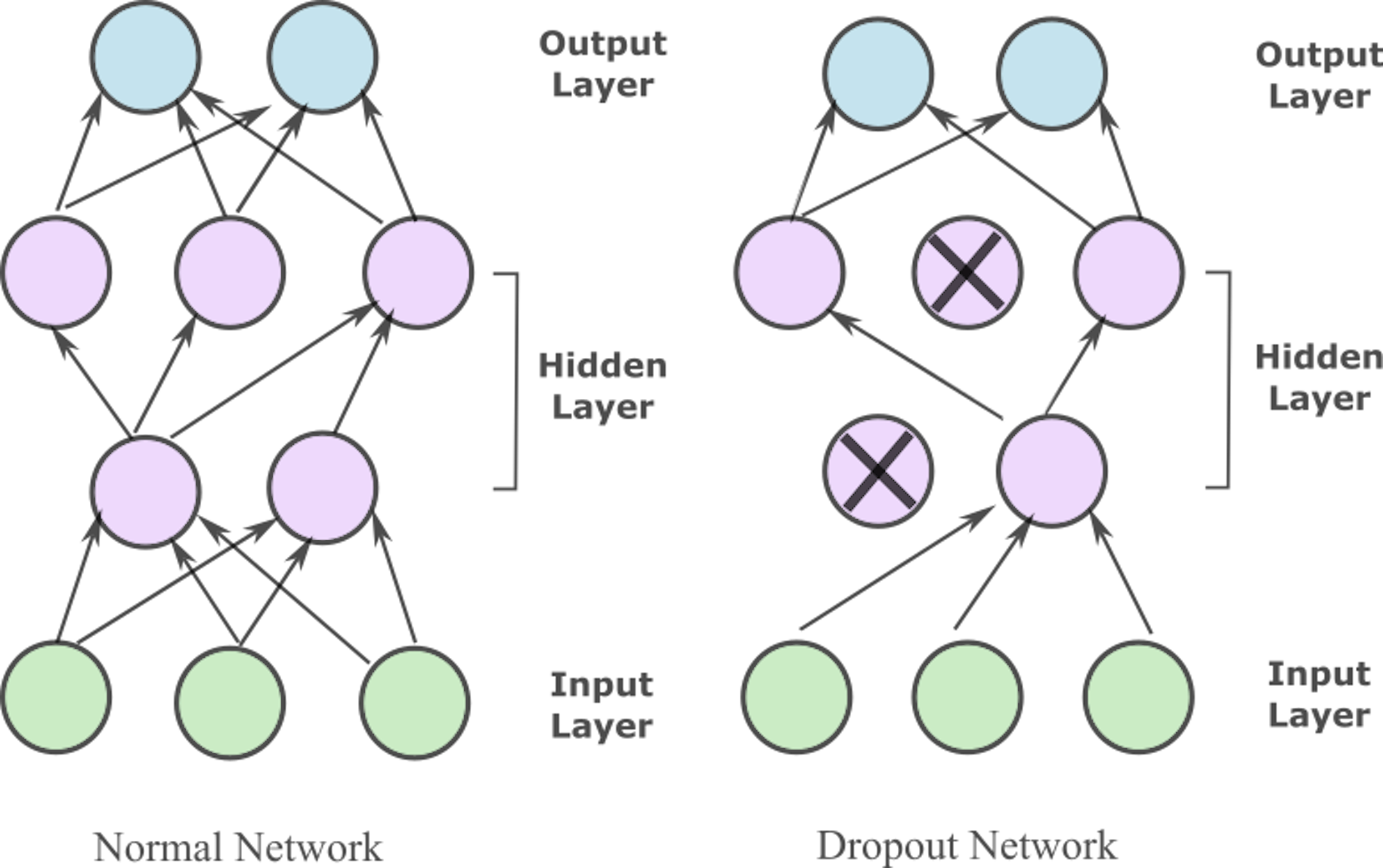}
    \caption{Dropout mechanism in a neural network.}
    \label{fig:dropout}
\end{figure}


\vspace{1em}
\begin{equation} 
Z_{w,t} \sim Bernoulli(p) \;  \forall  \: w \in \mathbf{W}
\label{ber}
\end{equation}
\vspace{1em}

In this case, \emph{T} samples are selected from the weight distribution (Equation \ref{lab1}) and used to perform \emph{T} stochastic forward passes with dropout, where the input is \textbf{X} and the output is \textbf{Y}. This allows the calculation of the expected value of the prediction Y and the variance of these predictions across iterations (Equations \ref{lab2} and \ref{lab3}). Thus, using T stochastic forward passes with different weight samples ${W_t}_{t=1}^T$, dropout acts as stochastic sampling. The observed variability in the T stochastic outputs primarily represents the model's inherent uncertainty or epistemic uncertainty. In other words, when predictions exhibit large variance, it suggests significant epistemic uncertainty in the model.

\begin{equation}
    \mathbf{W}_t  =  \text{train}(f; \mathbf{X, Y})
    \label{lab1}
\end{equation}

\begin{equation}
    \mathbb{E}(\hat{\mathbf{Y}}\mid\mathbf{X})  =  \frac{1}{T} \sum_{t=1}^{T}f(\mathbf{X}\mid{\mathbf{W}}_t)  
    \label{lab2}
\end{equation}

\begin{equation}
    Var(\hat{\mathbf{Y}} \mid \mathbf{X})  =  \frac{1}{T} \sum_{t=1}^{T} f(\mathbf{X})^{2} - \mathbb{E}(\hat{\mathbf{Y}}\mid\mathbf{X})^{2} 
    \label{lab3}
\end{equation}\\

\section{Related Work}
\label{sec:related}
Machine learning and deep learning have significantly benefited healthcare research. As medical data became more complex and accessible, traditional methods proved inadequate, leading to widespread integration of machine learning in healthcare analysis \cite{habehh2021machine}. Deep learning techniques emerged to handle complex, heterogeneous, and unstructured medical data from various sources, showing impressive results in biomedical research \cite{miotto2018deep}. Their applications include studying diseases, interpreting medical imaging, and forecasting health metrics. Consequently, cancer research has particularly benefited, with new studies exploring previously understudied areas like skin cancer \cite{kavitha2023detection}, colon cancer \cite{haq2023deep}, and viral cancer expressions \cite{elbasir2023deep}. Increased computing power has allowed deep learning to excel in computer vision, streamlining medicine and neuroimaging research \cite{sanchez2023machine}. These techniques are revolutionizing the diagnosis, treatment, and management of neurological conditions \cite{hussain2024mind}. Furthermore, some studies have used deep learning to generate synthetic medical data, addressing resource constraints \cite{hu2022detecting}. In some research studies, evidence has also been found supporting the clinical utility of these methods in decision support tools, laboratory diagnostics, and public health outbreak management \cite{theodosiou2023artificial, badawy2023healthcare}. As these technologies evolve, they promise to revolutionize healthcare practices and improve patient outcomes across various medical specialties.

Among the abstract data found in medical analysis, proteins stand out as both vital and prominent. These essential macromolecules are responsible for the structural and functional mechanisms of living organisms and have been extensively studied using various artificial intelligence techniques for protein classification, interaction prediction, and functional analysis \cite{ghosh2024matpip}. Such studies aim to classify and extract information from different protein families to serve various medical sectors. While these studies address specific aspects of protein analysis, recent breakthroughs have pushed the boundaries of AI applications in protein research further. Such as, in the study by Brandes et al. \mbox{\cite{brandes2022proteinbert}}, which introduced ProteinBERT, a deep learning model for protein sequences. Combining language modeling with Gene Ontology annotation prediction, this model achieves near state-of-the-art performance on multiple benchmarks while using a smaller and faster model than competing methods. Additionally, Senior et al. \mbox{\cite{senior2020improved}} developed AlphaFold, a groundbreaking protein structure prediction system that uses deep learning to predict highly accurate protein structures by leveraging genetic information and employing neural networks. These advancements demonstrate the growing capabilities of AI-driven approaches in protein research, offering promising tools for various applications in structural biology and drug discovery. More recently, transformer-based architectures have also been explored for extracting biological information directly from nucleotide and protein sequences. For example, Stefanini et al. employed the Perceiver architecture to predict gene and protein expression levels from DNA and protein sequences, demonstrating the effectiveness of attention-based models in sequence learning \cite{stefanini2023predicting}

Within this vast protein universe, phage structural proteins, also known as phage virion proteins (PVPs), represent another crucial subgroup. As essential components of bacteriophage virions, these proteins are associated with the most abundant biological entities on Earth—bacteriophages \cite{cobian2016viruses}. As high-throughput sequencing technologies advance, the need for robust computational frameworks to analyze and characterize newly discovered bacteriophages becomes increasingly important for future researches in this field \cite{nami2021application}. Based on recent literature reviews \cite{meng2020review, nami2021application}, the classification of phage virions using machine learning techniques has shown promising results. Commonly employed algorithms such as Naive Bayes (NB), Random Forest (RF), Support Vector Machine (SVM), and Scoring Card Method (SCM) have demonstrated satisfactory classification accuracy, typically ranging from 70\% to 80\% on test sets.
Building upon these foundational approaches, researchers have developed more sophisticated methods for phage virion classification. Seguritan et al. introduced iVirions \cite{seguritan2012artificial}, a deep learning approach leveraging amino acid frequency and protein isoelectric points for binary classification of major capsid and tail proteins. Similarly, Fang et al. developed VirionFinder \cite{fang2021virionfinder}, utilizing biochemical properties and one-hot encoding of sequences to identify complete and partial virion proteins via 1-D CNNs. Expanding on binary classification, Cantu et al. created PhaANN \cite{PhANN}, achieving 86.2\% test accuracy across multiple classes by integrating dipeptide and tripeptide composition features within artificial neural networks (ANNs). Building upon these advancements, Fang et al. introduced DeePVP \cite{fang2022deepvp}, employing one-hot encoding and 1-D CNNs for multiclass classification, demonstrating enhanced performance, particularly in binary classification tasks compared to PhaANN.

Diverging from prior methodologies, Shang et al. \cite{phaVIP} utilized an image-based FCGR encoding on sequences, employing a vision transformer model for classifying 8 subclasses, yielding superior results on standard evaluation metrics. Image representations of such 1-dimensional sequences have been widely adopted in literature for various protein-related tasks \cite{pang2017novel}. Additionally, Akbari et al. developed WalkIm \cite{akbari2022walkim}, an image-based encoding method enhancing biological sequence classification using CNNs with the DNA-walk algorithm. Collectively, these approaches demonstrate the versatility and effectiveness of image-based representations in protein sequence analysis and classification tasks across various domains.

\section{Methodology}
\label{sec:method}
The following sections detail our methodology, covering the dataset description, encoding algorithm, pre-trained classifiers for phage classification, and the use of MCD for prediction uncertainty estimation. Figure \ref{fig:flow} illustrates the workflow, showcasing key stages of data processing, feature extraction, classification, and uncertainty quantification.

\begin{figure*}
    \centering
    \includegraphics[width=\linewidth]{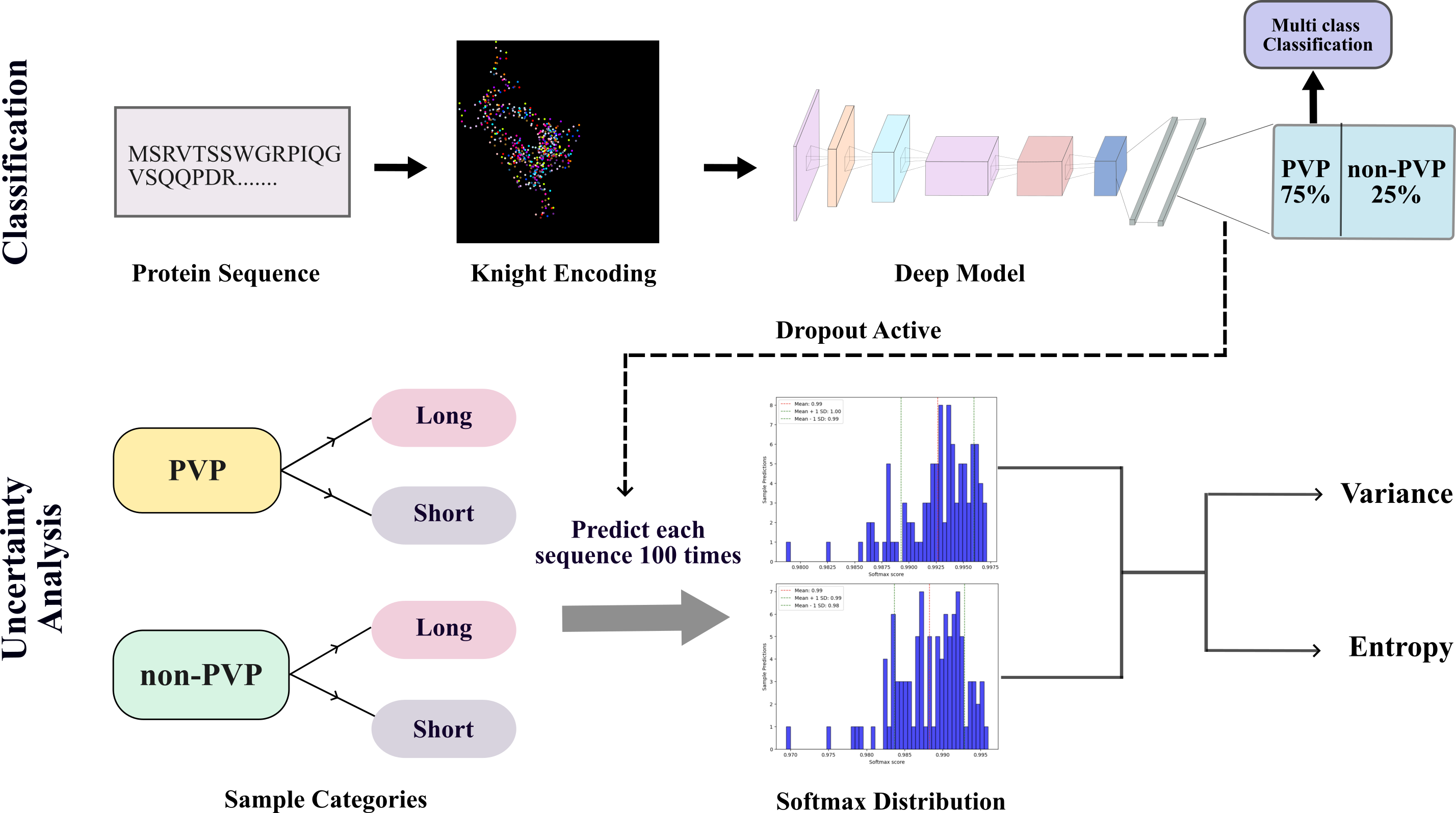}
    \caption{PVP classification and uncertainty analysis framework}
    \label{fig:flow}
\end{figure*}

\subsection{Dataset Description}

\begin{figure}[H]
    \centering
    \includegraphics[width=\columnwidth]{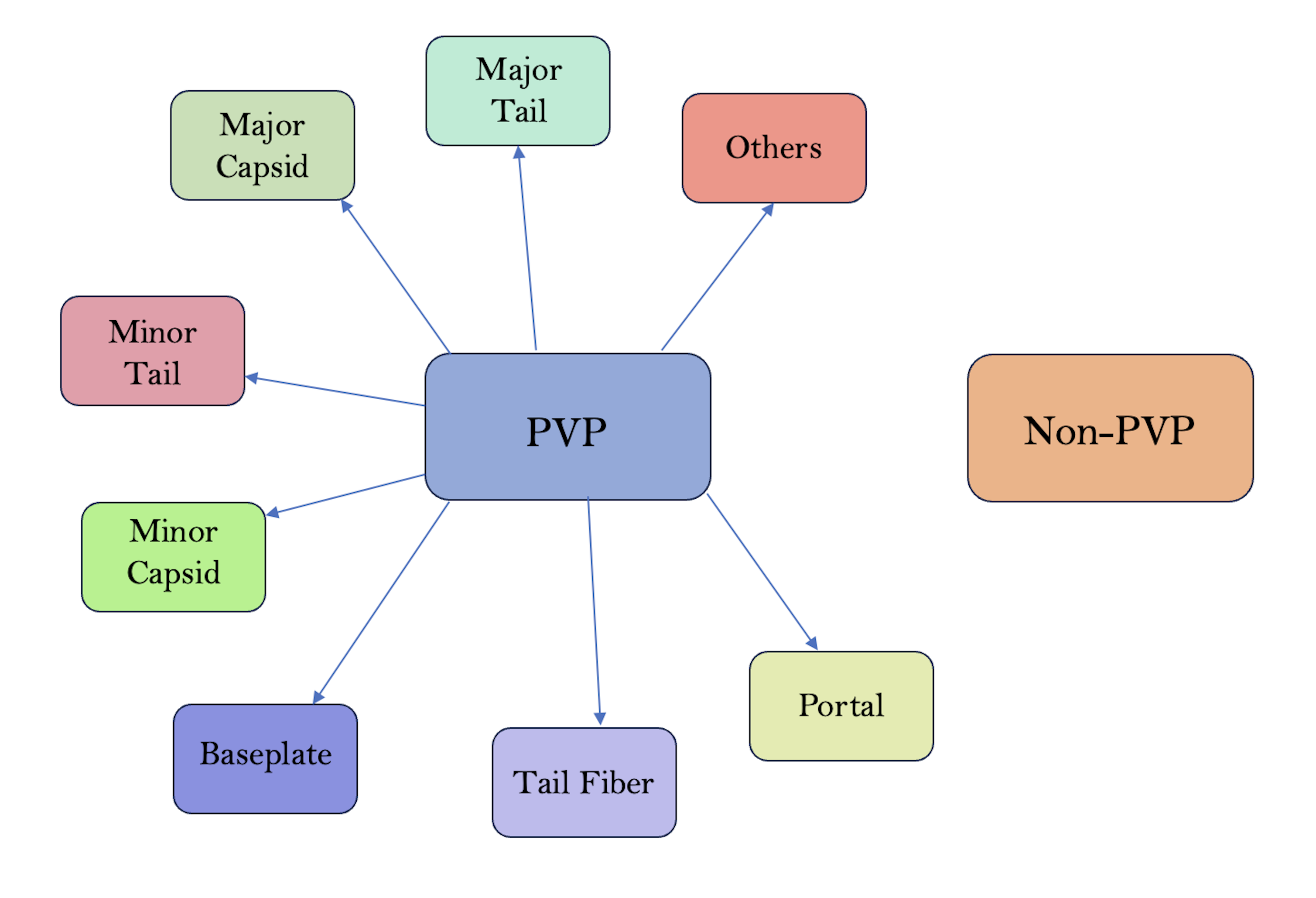}
    \caption{Illustration of dataset classes.}
    \label{fig:types}
\end{figure}

For this study, we utilized a benchmark dataset \cite{phaVIP}. Sequences with low-confidence labels, which introduced ambiguity, were removed from the dataset. Subsequently, a search was conducted using diverse keywords to identify and extract structural proteins from the remaining sequences. Non-structural proteins were collected by identifying enzymes with names ending in the suffix `ase'. The CD-hit algorithm was then applied to identify clusters of sequences with a similarity threshold of 90\%. The longest sequence within each cluster was chosen as the representative sequence. The final database comprises a total of \textbf{35,213} PVP sequences and \textbf{46,883} non-PVP sequences of varying lengths. The PVP sequences are further categorized into eight groups, curated for multi-class classification purposes. Figure \ref{fig:types} illustrates the overall class breakdown, while Table \ref{mult} provides a summary of the multiclass data counts.

\subsection{`Knight' Encoding}

Protein sequences lack image-based encoding compared to DNA sequences. For DNA sequences, the random walk or DNA walk method has been widely used for encoding for sequence analysis \cite{berger2004visualization}, achieving high prediction accuracy. The study \cite{akbari2022walkim} utilizes DNA-walk to classify viral genomes using CNNs, suggesting it as an alternative to k-mer-based encodings like FCGR, which may lose spatial information of encoded sequences. They propose its usability for other sequence types such as RNA and proteins, although no such implementation has been reported. Based on this, we introduce a novel walk-based encoding technique specifically for protein sequences, aiming to evaluate its effectiveness in characterizing phage structural proteins.
This encoding method employs polar coordinates to interpret protein sequences. Each amino acid is assigned an angle value, totaling 360 degrees, representing its directional movement. Unique color-coded markers or circular points depict the amino acids, positioned on the image based on their angle and distance of movement (radius). Equations \ref{1e} and \ref{2e} are used to calculate the x and y direction displacements for encoding residues from their current positions.
\begin{equation}
    x = r \times \cos({\theta})
    \label{1e}
\end{equation}
\begin{equation}
    y = r \times \sin({\theta})
    \label{2e}
\end{equation}
\vspace{0.5cm}

The entire encoding algorithm can be divided into three major parts:

\begin{enumerate}
    \item 
    \textbf{Definitions} \\
    A list is defined that includes the letter representation of each amino acid, along with a complementary dictionary that maps each amino acid to a corresponding color. These are subsequently utilized for the computation of angles and placement of points.

      \begin{table}[htbp]
        \caption{PVP multi-class data entries}\label{mult}%
         \centering
         \begin{tabular}{l c}
            \hline 
              \textbf{PVP Classes} & \textbf{Number of entries}\\
              \hline
               Baseplate &  3362\\
              \hline 
               Portal & 2770\\
               \hline
               Tail Fiber & 2305\\
               \hline
               Major Capsid & 2443\\
               \hline
               Minor Capsid & 398\\
               \hline
               Major Tail & 5083\\
               \hline
               Minor Tail & 1458\\
               \hline
               Others & 17385\\
               \hline
         \end{tabular}
     \end{table}
\FloatBarrier

List of amino acids for the encoding,

\begin{minipage}{\linewidth}
\begin{lstlisting}
self.amino_acids = ['A', 'C', 'D', 'E', 'F', 'G',
                    'H', 'I', 'K', 'L', 'M', 'N',
                    'P', 'Q', 'R', 'S', 'T', 'V',
                    'W', 'Y']
\end{lstlisting}
\end{minipage}

Amino acid to color mapping,

\begin{minipage}{\linewidth}
\begin{lstlisting}
self.colors = {
    'A': (255, 0, 0),     'C': (255, 255, 0),
    'D': (0, 234, 255),   'E': (170, 0, 255),
    'F': (255, 127, 0),   'G': (191, 255, 0),
    'H': (0, 149, 255),   'I': (255, 0, 170),
    'K': (237, 185, 185), 'L': (185, 215, 237),
    'M': (231, 233, 185), 'N': (220, 185, 237),
    'P': (185, 237, 224), 'Q': (143, 35, 35),
    'R': (35, 98, 143),   'S': (143, 106, 35),
    'T': (107, 35, 143),  'V': (115, 237, 155),
    'W': (204, 204, 204), 'Y': (0, 64, 255)
}
\end{lstlisting}
\end{minipage}

\par
\item 
\textbf{Angle Calculation}
\begin{enumerate}
\item We have employed the structure of a 20-sided polygon called `Icosagon' to represent the 20 amino acids. The amino acids are evenly distributed throughout the 20 vertices of the Icosagon. Each point is separated from the others by a constant angle of $18^{\circ}$ $(360^{\circ} / 20 = 18^{\circ})$, as seen in Figure \ref{fig:icosagon}.

\begin{figure}[htbp]
\centering
\includegraphics[scale=0.45]{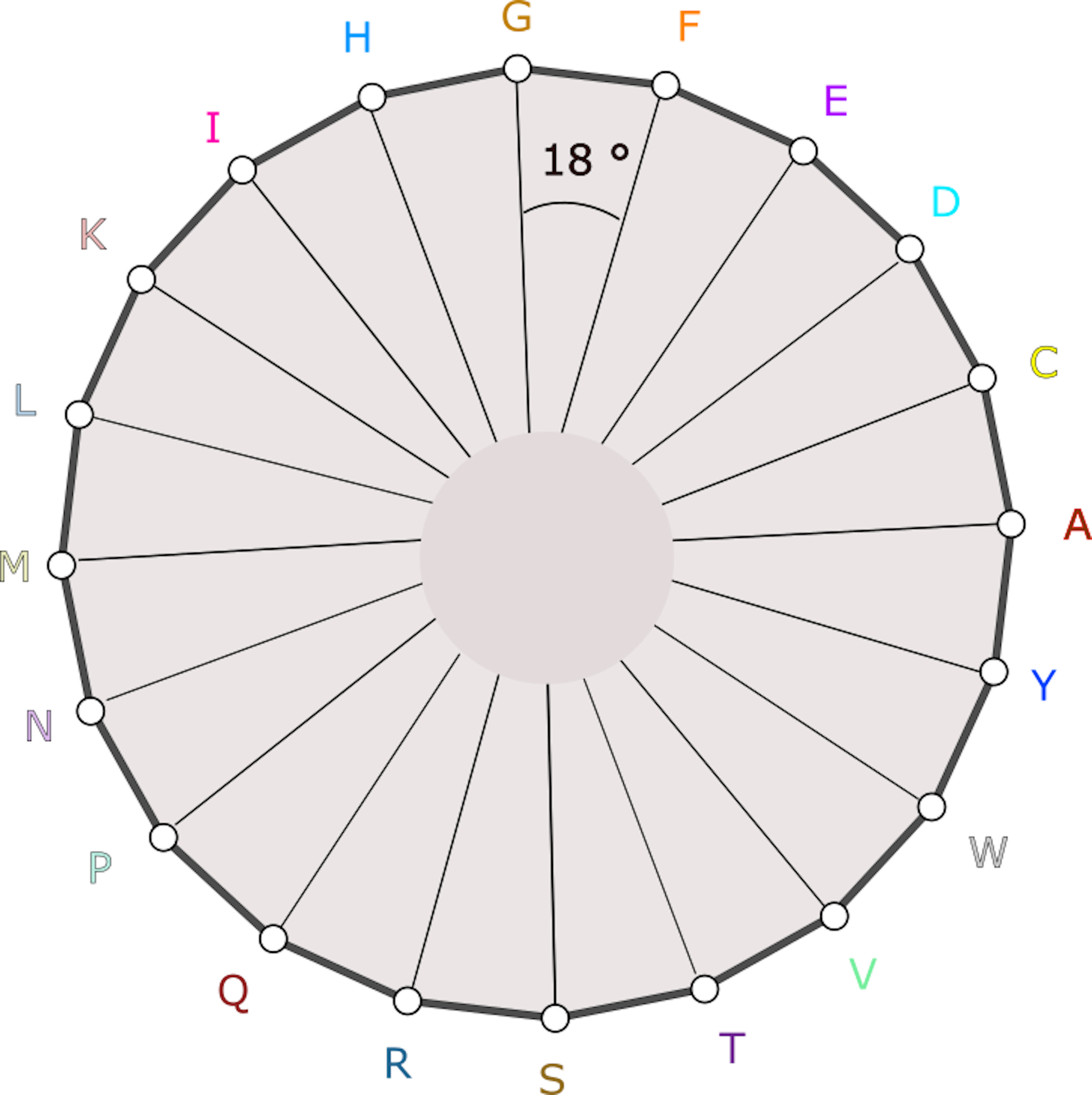}
\caption{Icosagon with color-coded vertices and 18° angular assignments for amino acids}
\label{fig:icosagon}
\end{figure}

\item The angles for each letter in the amino acid list are determined based on the index position of the amino acids in the list mentioned previously. For instance, C is at index 1, and G is at index 5.
\\\\So for `C', the associated angle is,
$$\text{index of C} = 1$$
$$\theta = 1 \times 18 ^{\circ} = 18 ^{\circ}$$
\\ Similarly, for `G', the angle will be,
$$\text{index of G} = 5$$
$$\theta = 5 \times 18 ^{\circ} = 90 ^{\circ}$$
As we shall be using polar coordinate formulas in our study, the angles are converted from degrees to radians.
\end{enumerate}
\vspace{0.5cm}
\item \textbf{Placement}
\vspace{0.1cm}
\begin{enumerate}
\item If the dimensions of the image are defined as $M \times M$, then the encoding will start from the origin, which is located at the coordinates $(x, y) = (M/2, M/2)$, or in other words, from the middle of the image.
\item For the first amino acid (or the first letter of our sequence), $(x, y) = (M/2, M/2)$ or the center of our image will act as the starting point, i.e., current coordinates.
\item Using the fixed radius and corresponding angle of an amino acid, the horizontal and vertical displacement $(x\prime, y\prime)$ of the encoding is determined using equations \ref{1e} and \ref{2e}, relative to the current coordinates $(x, y)$, as shown in equations \ref{x} and \ref{y}.
\begin{equation}
x = x + x\prime
\label{x}
\end{equation}
\begin{equation}
y = y - y\prime
\label{y}
\end{equation}
A circular point will be positioned at the coordinates $(x, y)$ to represent the amino acid being encoded, with its corresponding color. The polar coordinate radius and point size remain consistent at 15 and 2, respectively, for all amino acids.
\item Subsequently, the next character of the protein sequence will utilize the coordinates $(x, y)$ of the previous character as its starting point. A new horizontal and vertical shift $(x\prime, y\prime)$ will be generated based on the amino acid's associated angle. Using these new shift values $(x\prime, y\prime)$, a new set of coordinates $(x, y)$ are calculated for placing the current amino acid character.
\item The remaining characters in the sequence will continue to follow the previously specified technique until the entire sequence is encoded, as shown in Figures \ref{fig:fig1} and \ref{fig:fig2}. If a point hits the boundary of the image during encoding, it will begin encoding from the center of the image at $(x, y) = (M/2, M/2)$.
\end{enumerate}
\end{enumerate}

\begin{figure}[H]
\centering
\includegraphics[scale=0.4]{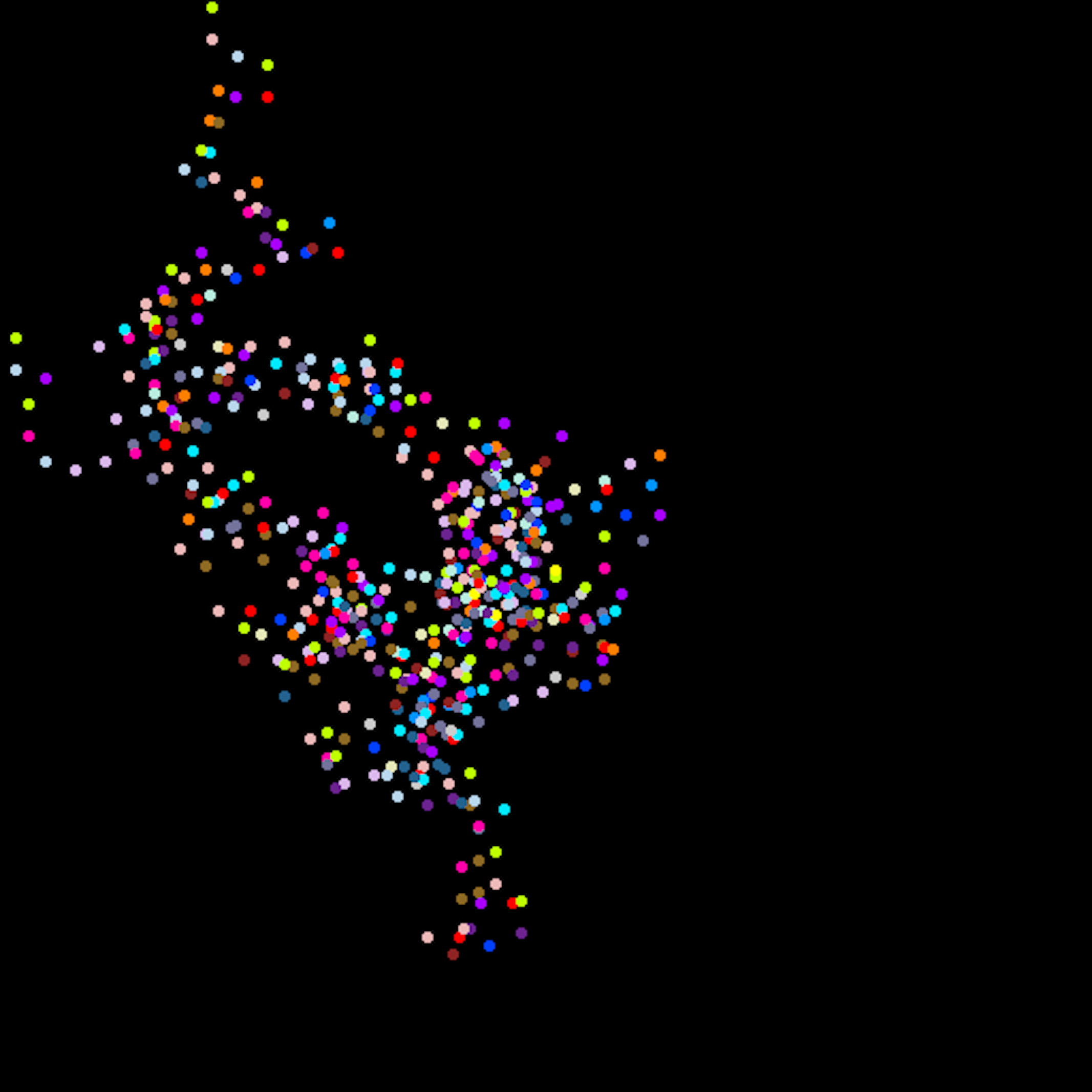}
\caption{PVP(YP\_009900749.1 minor capsid protein [Lactococcus phage 62503]}
\label{fig:fig1}
\end{figure}

\begin{figure}[H]
\centering
\includegraphics[scale=0.4]{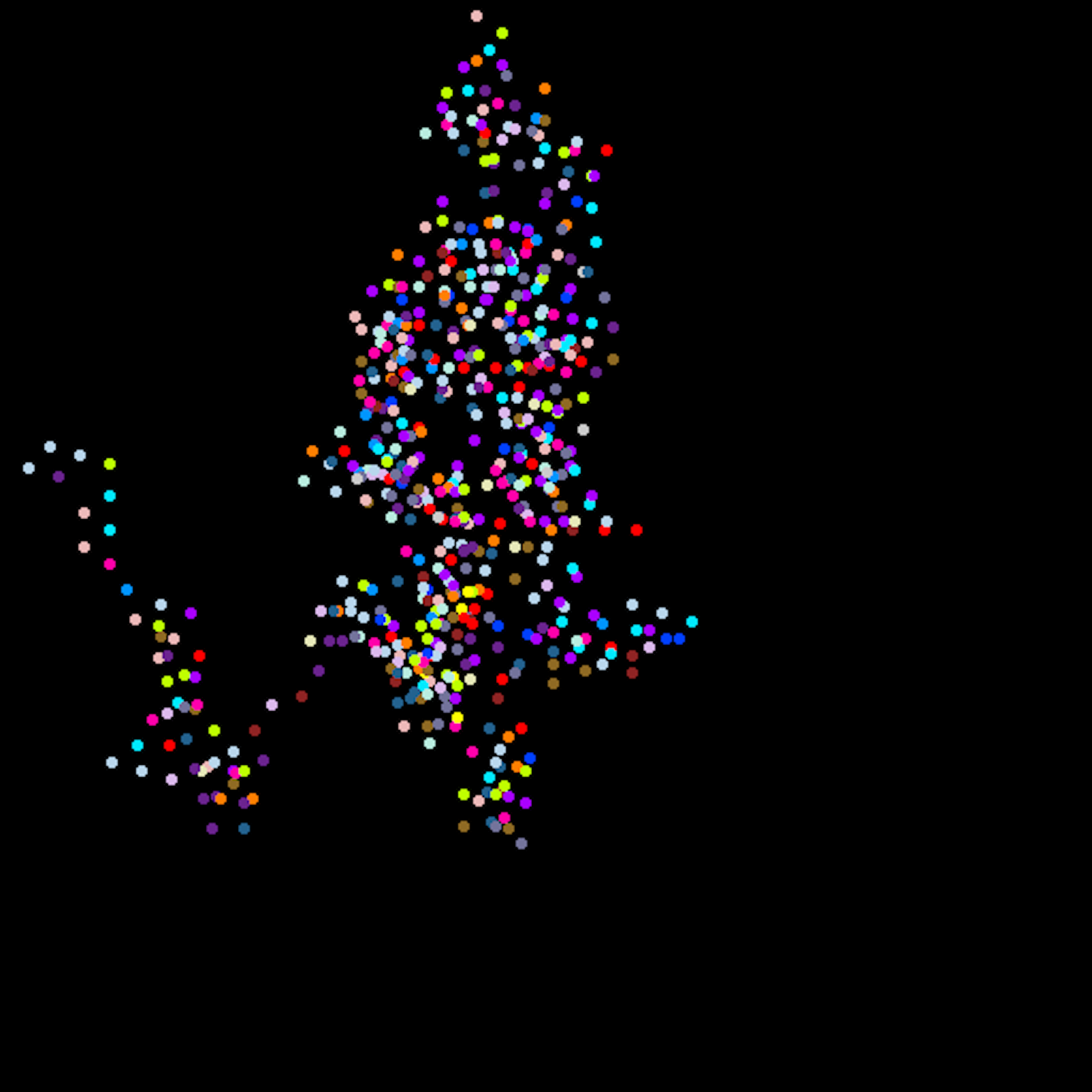}
\caption{non-PVP(YP\_009847768.1 (NAD(+)) DNA ligase [Vibrio phage USC-1])}
\label{fig:fig2}
\end{figure}
\FloatBarrier

In summary the Knight encoding algorithm (\ref{algo1}) is given as follows:

\begin{algorithm}
\caption{Encoding a Protein Sequence into an $M \times M$ Image}
\label{algo1}
\DontPrintSemicolon
\KwIn{Protein sequence $S$ of length $N$; image size $M$; step radius $r = 15$;
      residue ordering $amino\_acids$ ($20$ residues); color map $colors$}
\KwOut{Encoded RGB image $I$ of size $M \times M$}
$(x, y) \leftarrow \big(\lfloor M/2 \rfloor,\ \lfloor M/2 \rfloor\big)$\;
\For{$i \leftarrow 0$ \KwTo $N-1$}{
    $residue \leftarrow S[i]$\;
    $\theta \leftarrow 18^{\circ} \times amino\_acids.\mathrm{index}(residue)$\;
    $x \leftarrow x + \mathrm{trunc}(r\cos\theta)$\;
    $y \leftarrow y - \mathrm{trunc}(r\sin\theta)$\;
    \If{$x \notin (0, M)$ \textbf{\emph{or}} $y \notin (0, M)$}{
        $(x, y) \leftarrow \big(\lfloor M/2 \rfloor,\ \lfloor M/2 \rfloor\big)$\;
    }
    $color \leftarrow colors[residue]$\;
    Plot a point of size $2$ at $(x, y)$ in $I$ using $color$\;
}
\end{algorithm}

\subsection{Pre-trained CNN Models}

CNNs have demonstrated remarkable efficacy across various domains of computer vision tasks. To extend the utility of such powerful models to fields like medicine and bioinformatics, where large datasets are often limited, transfer learning is employed to train models \cite{salama2021deep}. Besides circumventing the need for extensive datasets, transfer learning facilitates the construction of models in a resource-efficient and time-saving manner \cite{rawat2017deep}. Pre-trained CNNs have exhibited competitive performance compared to popular pre-trained transformer models, further highlighting their relevance in image classification tasks \cite{tan2018identifying}. Therefore, for our study, we have utilized existing pre-trained CNN variants such as GoogLeNet, EfficientNet, and MobileNet, focusing on models with fewer parameters yet capable of achieving superior results.

\subsection{Estimating Uncertainty using MCD}

To assess uncertainty in binary classification, we activate the model's dropout layer with a 0.2\% dropout rate during the test phase. Consequently, we develop a method to classify data into distinct groups based on sequence properties to analyze uncertainty for each category. Unlike text data, protein sequences do not reveal unique variability apart from class labels. The only noticeable diversity is sequence length, ranging from hundreds to thousands of residues (Figures \ref{fig:pvp} and \ref{fig:non-pvp}). Therefore, investigating the model's robustness to sequence length can determine its vulnerability to uncertainty. The training data were divided based on an equilibrium sequence length $\delta$, ensuring nearly equal data with lengths less than or equal to $\delta$ and greater than $\delta$ to mitigate length bias (Table \ref{tab1}).

\begin{table}[htbp]
\caption{Data Distribution for PVP and non-PVP $\delta$-values}
\label{tab1}
\centering
\small
\begin{tabular*}{\columnwidth}{@{\extracolsep{\fill}}lccc@{}}
\toprule
\textbf{Class} &
\makecell{\textbf{Sequences}\\\textbf{$< \delta$}} &
\makecell{\textbf{Sequences}\\\textbf{$> \delta$}} &
\makecell{\textbf{Total}\\\textbf{Sequences}} \\
\midrule
PVP ($\delta = 350$)      & 12589 & 12060 & 24649 \\
\hline
non-PVP ($\delta = 275$)  & 16502 & 16316 & 32818 \\
\bottomrule
\end{tabular*}
\end{table}

\begin{figure*}
    \centering
    \includegraphics[width=\linewidth]{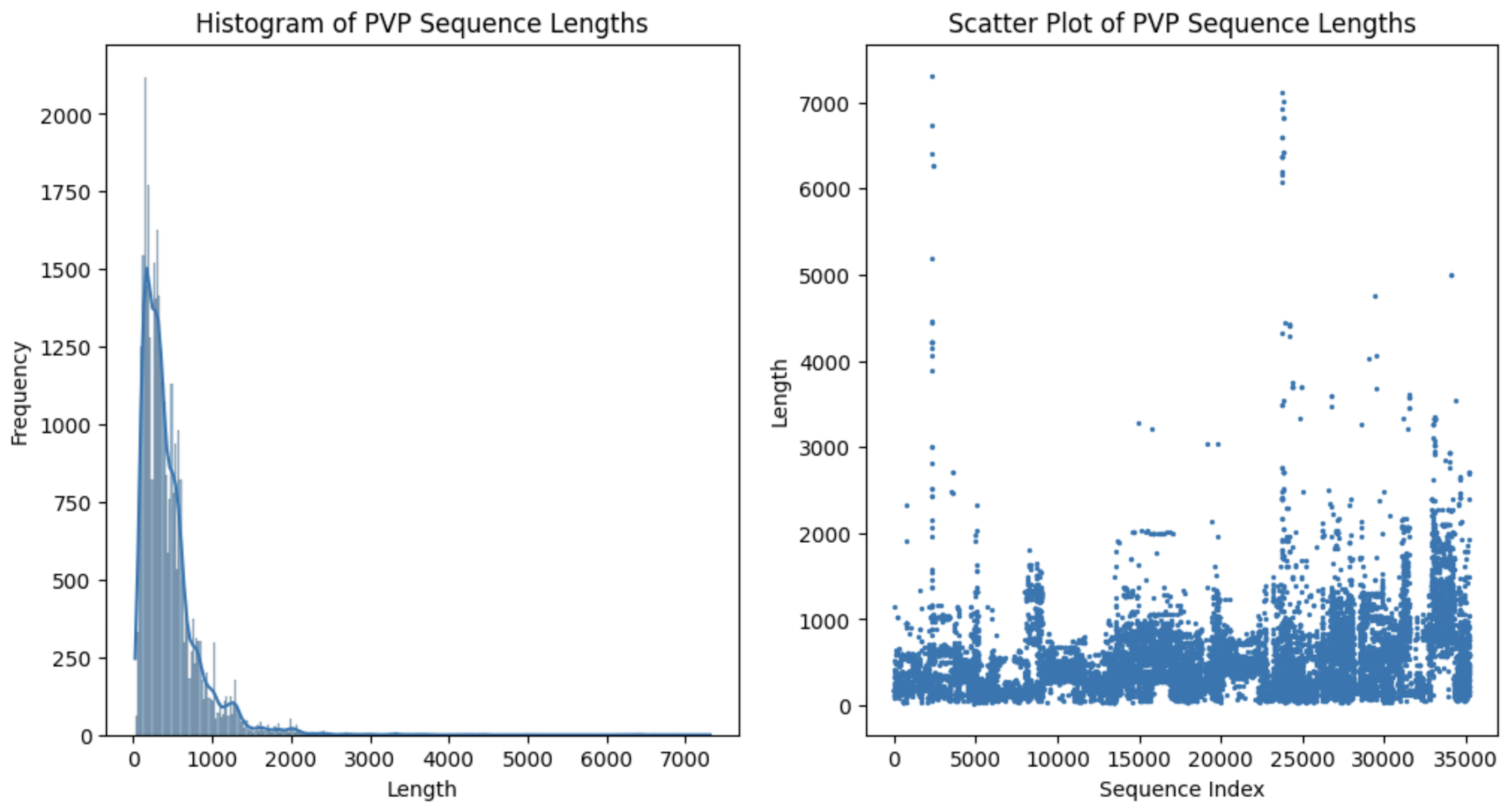}
    \caption{Length Distribution of PVP sequences}
    \label{fig:pvp}
\end{figure*}

\begin{figure*}
    \centering
    \includegraphics[width=\linewidth]{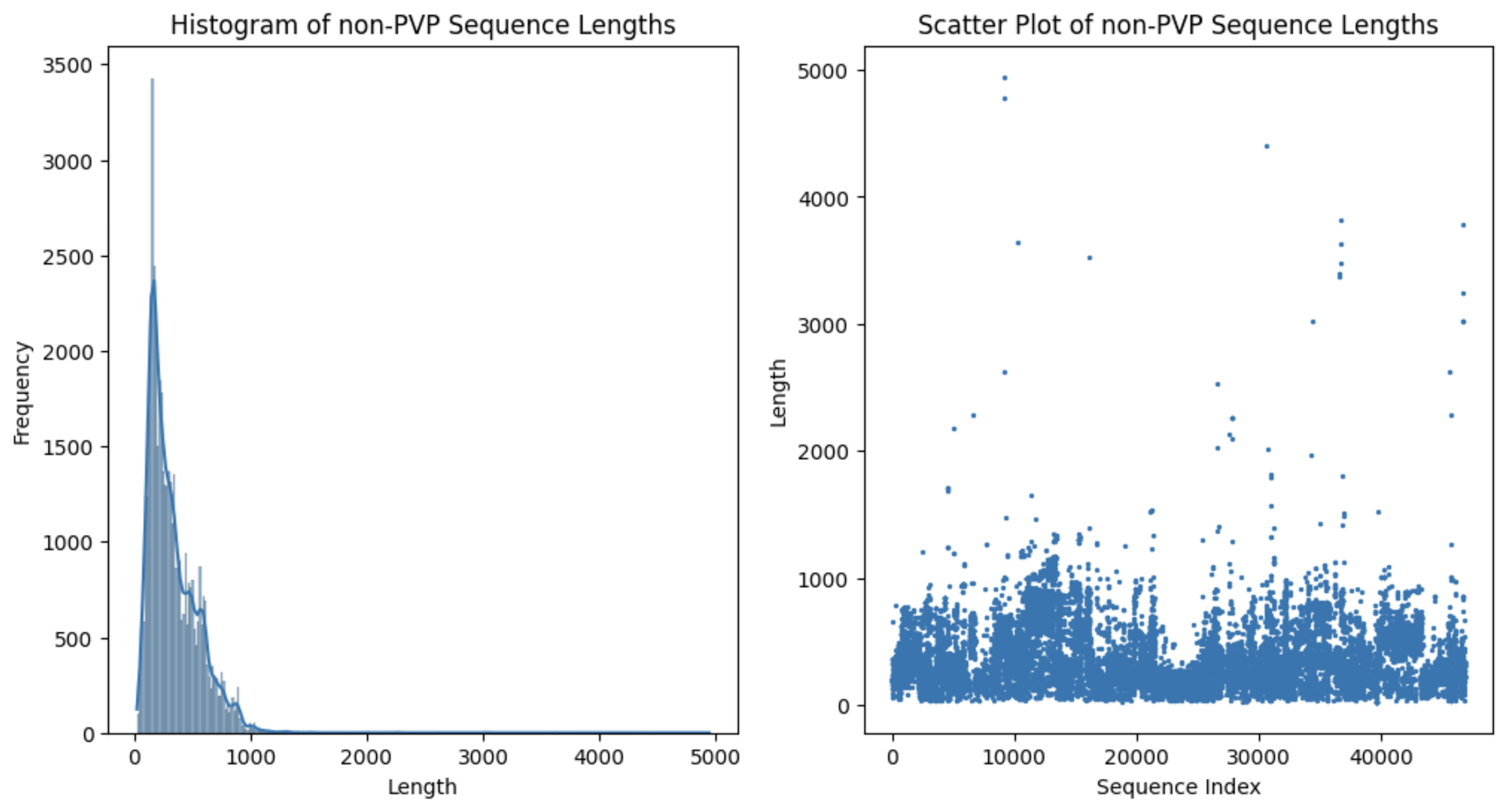}
    \caption{Length Distribution of non-PVP sequences}
    \label{fig:non-pvp}
\end{figure*}

The data are separated into four categories: PVP ($<$ 350), PVP ($>$ 350), non-PVP ($<$ 275), and non-PVP ($>$ 275). For each category, 100 randomly chosen sequences are predicted 100 times using the dropout model, providing a prediction distribution. The predictions are passed through the softmax activation, transforming raw logits into probabilities. The mean and variance of these softmax probabilities from each category are determined to compare uncertainty across categories. As shown in Figures \ref{fig:soft1} and \ref{fig:soft2} for a sample, we expect a probability distribution, with the x-axis representing softmax scores and the y-axis indicating the number of predictions for each score.

\begin{figure}[htbp]
\centering
\includegraphics[width=\columnwidth]{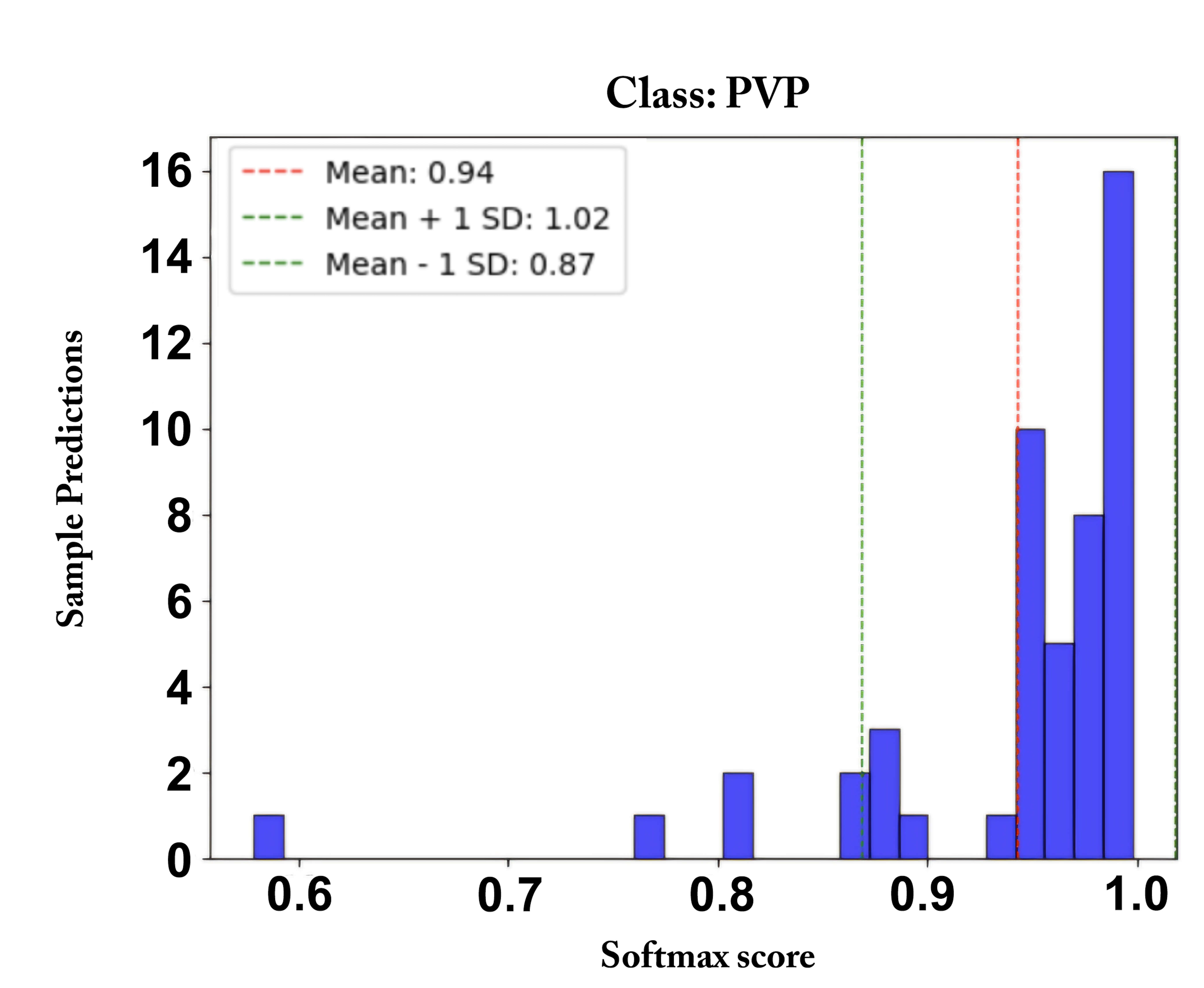}
\caption{Softmax Distribution of PVP}
\label{fig:soft1}
\end{figure}

\begin{figure}[htbp]
\centering
\includegraphics[width=\columnwidth]{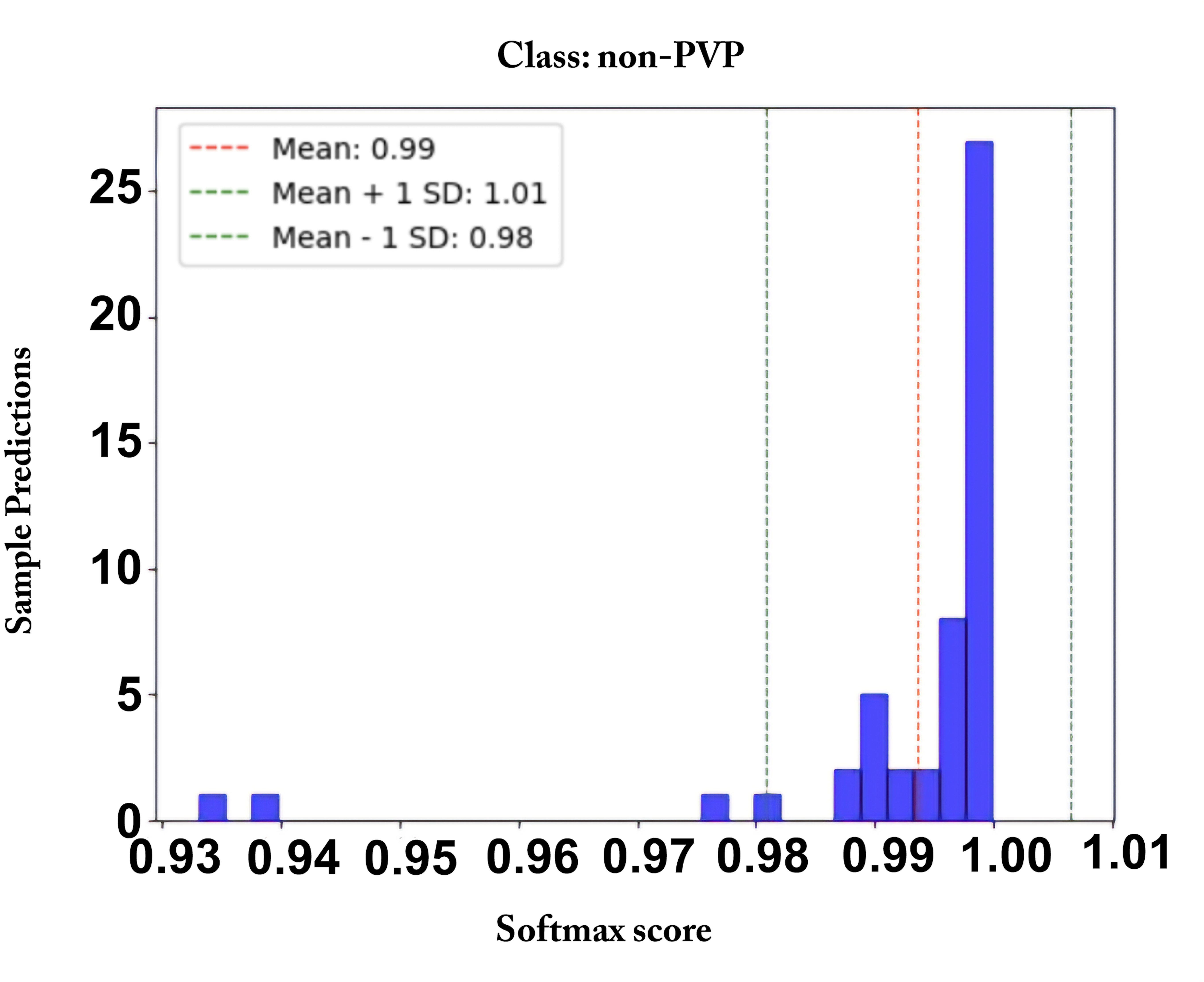}
\caption{Softmax Distribution of non-PVP}
\label{fig:soft2}
\end{figure}

Typically, correct predictions should cluster on the right side of the axis, as a softmax score greater than 0.5 indicates the data belongs to that class. Conversely, predictions toward the left denote incorrect predictions. The histogram's dispersion, or prediction variance, signifies the model's confidence level. Reduced variance indicates higher confidence, while higher variance suggests greater prediction uncertainty for that specific data.

\section{Experimental Analysis}
\label{sec:experiments}
\subsection{Experimental configuration}

The training and testing procedures for the experiments were developed and implemented using a comprehensive suite of Python libraries, including TensorFlow, Keras, and PyTorch. The computational resources utilized for model training and evaluation consisted of a high-performance computer equipped with an AMD Ryzen 7 3700X 8-core CPU, 16GB of DDR4 RAM, and an NVIDIA GeForce RTX 3060 Ti graphics processing unit. The experiments were conducted on the Windows 11 Pro operating system, leveraging the Python 3.11 interpreter as the primary programming environment.

\subsection{Training and Evaluation}

To evaluate the pre-trained CNN models, each was fine-tuned on the encoded image dataset for 25 epochs, with a batch size of 32. Accuracy was the primary metric, and `binary\_crossentropy' was the optimization loss function. Accuracy was considered the primary metric, as correctly identifying positive and negative samples is equally important. Additionally, recall/sensitivity, specificity, precision, and F1-score were calculated for a comprehensive performance overview.
By evaluating diverse metrics, a thorough understanding of the CNN models' performance was obtained, facilitating optimal model selection. These additional metrics were derived from the true positive (TP), true negative (TN), false positive (FP), and false negative (FN) predictions:

\begin{equation}
     Accuracy = \frac{TP + TN}{TP + TN + FP + FN}
\end{equation}

\begin{equation}
     Precision = \frac{TP}{TP + FP}
\end{equation}

\begin{equation}
     Recall/Sensitivity = \frac{TP}{TP + FN}
\end{equation}

\begin{equation}
    Specificity = \frac{TN}{TN + FP}
\end{equation}

\begin{equation}
    F_1 = \frac{2 \times Precision \times Recall}{Precision + Recall}
\end{equation}

\subsection{Experimental Results}

\subsubsection{Model Predictions}

As outlined in the literature review section, the PVP classification task encompasses both binary and multiclass categorization challenges. To comprehensively evaluate the efficacy of our proposed encoding approach, we employed deep learning techniques, which have been demonstrated to outperform traditional machine learning methods in terms of prediction accuracy and feature extraction simplicity. Rather than constructing an entirely new model from scratch, we leveraged several high-performing pre-trained CNNs and fine-tuned them on our
\begin{figure*}
    \centering
    \includegraphics[width=0.9\textwidth, height=0.3\textheight, keepaspectratio]{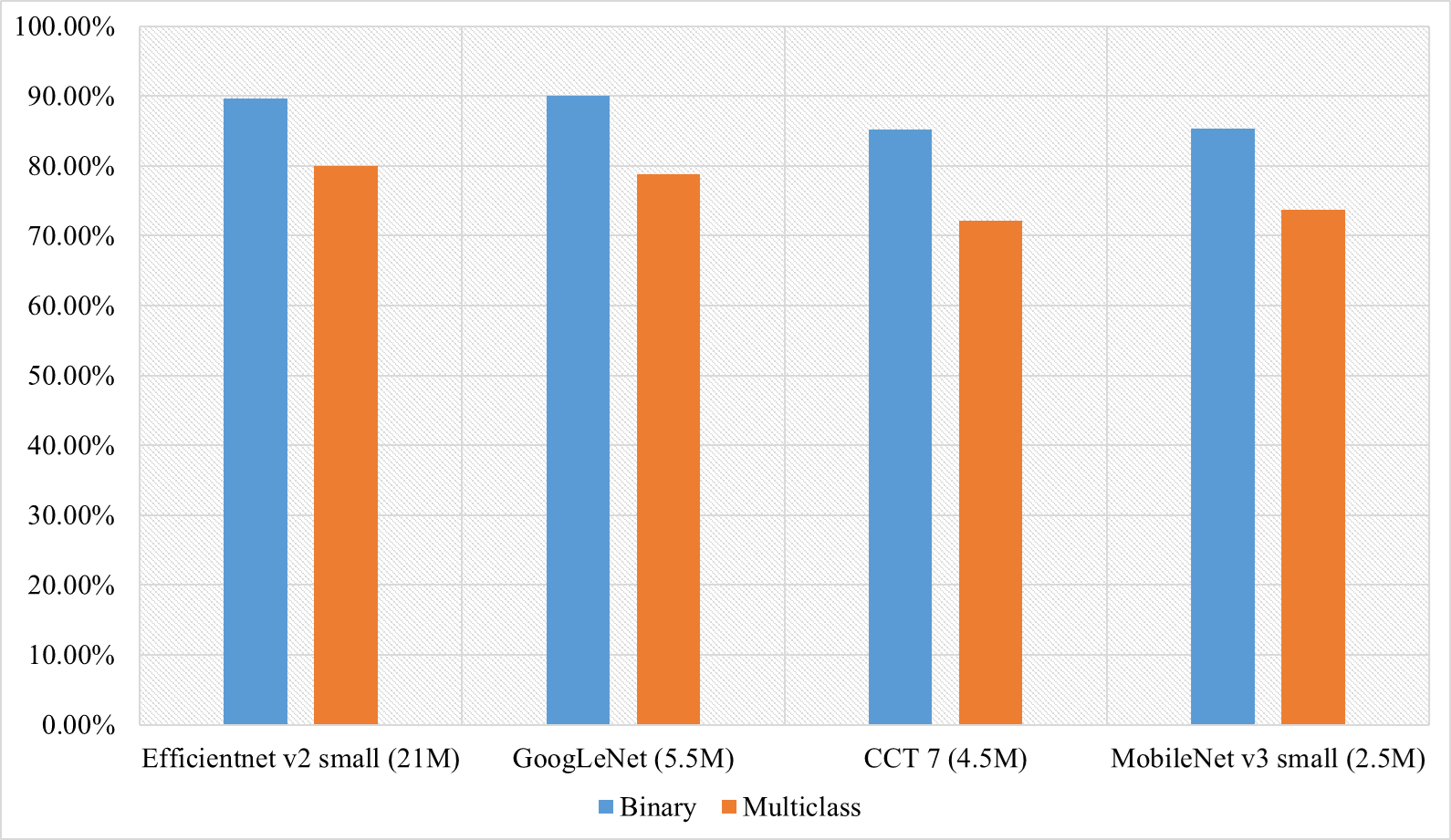}
    \caption{Model comparison}
    \label{fig:model_comparison}
\end{figure*}
dataset. Our particular interest lies in assessing the encoding efficiency under resource-constrained conditions, which motivated the selection of the specific CNN models employed in this study. To this end, we conducted a comparative analysis of various pre-trained CNN architectures to identify the most suitable model for our classification task.

Several pre-trained models from the PyTorch library were utilized. Among these, the top four CNN architectures that demonstrated satisfactory performance on our dataset, along with their respective parameters (provided in brackets) and prediction accuracy are illustrated in Figure \ref{fig:model_comparison}. All these CNNs exhibited competitive performance in terms of binary and multiclass classification, with EfficientNet v2 yielding the best results, surpassing GoogLeNet by a slight margin. However, GoogLeNet gained an advantage due to its significantly smaller parameter count, roughly one-fourth that of EfficientNet v2. Ultimately, the GoogLeNet architecture achieved an optimal trade-off between computational efficiency and classification quality.


Based on the evaluation results, the GoogLeNet architecture emerged as the most promising candidate model for further analysis. This analysis includes comparative assessments with previous works and in-depth uncertainty quantification. Table \ref{GoogleNet} presents the GoogLeNet model's performance across various evaluation metrics, while Table \ref{Hyperparameter} outlines the optimal hyperparameters used in training. Here, ``Image Size'' refers to the original size of the images generated from protein sequences after encoding, while ``Image Input Size'' refers to the final resized dimensions of the images as fed into the classification models'. The layer-wise parameters of GoogLeNet are the same as those provided in \cite{website}.
Evaluating on the test set, our GoogLeNet-based method achieved an accuracy of approximately 89\%, which is competitive with state-of-the-art methods. Notably, the precision, sensitivity, and F1-score of our GoogLeNet model surpassed most previous methods, as shown in Table \ref{canalysis_binary}. Considering the results of binary classification, our study achieved sensitivity and specificity of 91.1\% and 92.7\%, respectively. For comparison, we also tested EfficientNet V2, which achieved a slightly higher accuracy of around 91.2\%. Table \ref{canalysis_binary} provides a comprehensive comparison of the results of our GoogLeNet-based approach with existing state-of-the-art tools for the PVP classification task.

\begin{table}[htbp]
\caption{Evaluation results of the optimal model (GoogLeNet)}
\label{GoogleNet}
\centering
\begin{tabular*}{\columnwidth}{@{\extracolsep{\fill}}lcccc@{}}
\toprule
\textbf{Task} & \textbf{Params} & \textbf{Accuracy} & \textbf{F1 Score} & \textbf{Recall} \\
\midrule
Binary      & 5.5M & 88.60\% & 89.87\% & 88.54\% \\
Multiclass  & 5.5M & 76.37\% & 76.37\% & 76.37\% \\
\bottomrule
\end{tabular*}
\end{table}

\begin{table}[htbp]
\caption{Optimal hyperparameters used for model training}
\label{Hyperparameter}
\centering
\small
\begin{tabular*}{\columnwidth}{@{\extracolsep{\fill}}ccccc@{}}
\toprule
\makecell{\textbf{Image}\\\textbf{Size}} &
\makecell{\textbf{Input}\\\textbf{Size}} &
\textbf{Epochs} &
\makecell{\textbf{Batch}\\\textbf{Size}} &
\makecell{\textbf{Learning}\\\textbf{Rate}} \\
\midrule
$512 \times 512$ & $224 \times 224$ & 25 & 32 & 0.001 \\
\bottomrule
\end{tabular*}
\end{table}

\begin{table*}[t]
\caption{Comparative analysis for binary classification}
\label{canalysis_binary}
\centering
\small
\setlength{\tabcolsep}{6pt}
\renewcommand{\arraystretch}{1.3}

\begin{tabular*}{\textwidth}{@{\extracolsep{\fill}}lcccccc@{}}
\toprule
&
\multicolumn{3}{c}{\textbf{PVP}} &
\multicolumn{3}{c}{\textbf{non-PVP}} \\
\cmidrule(lr){2-4}\cmidrule(l){5-7}
\textbf{Work} &
\textbf{Precision} &
\textbf{Recall} &
\textbf{F1-score} &
\textbf{Precision} &
\textbf{Recall} &
\textbf{F1-score} \\
\midrule
PhANN \cite{PhANN}                          & 0.81 & 0.87 & 0.84 & 0.86 & 0.80 & 0.83 \\
DeePVP \cite{fang2022deepvp}                & 0.99 & 0.51 & 0.68 & 0.67 & 1.00 & 0.80 \\
VirionFinder \cite{seguritan2012artificial} & 0.77 & 0.92 & 0.84 & 0.90 & 0.73 & 0.81 \\
Meta-iPVP \cite{charoenkwan2020meta}        & 0.76 & 0.84 & 0.80 & 0.82 & 0.73 & 0.77 \\
PVPred-SCM \cite{charoenkwan2020pvpred}     & 0.78 & 0.57 & 0.66 & 0.66 & 0.84 & 0.74 \\
PhaVIP \cite{phaVIP}                        & 0.93 & 0.95 & 0.94 & 0.95 & 0.93 & 0.94 \\
\textbf{ProteoKnight}                       & \textbf{0.83} & \textbf{0.89} & \textbf{0.90} &
\textbf{0.94} & \textbf{0.89} & \textbf{0.90} \\
\bottomrule
\end{tabular*}
\end{table*}

\subsubsection{Uncertainty Analysis}
Uncertainty estimations of deep learning models serve as a measure of the reliability of their predictions. As stated in the methodology, the variance of the prediction distribution during dropout passes is regarded as a standard uncertainty quantification metric. Entropy (H) is another commonly utilized metric in prior research focused on uncertainty quantification, as demonstrated by Milanés et al. \cite{milanes2021monte} and Kendall et al. \cite{kendall2017uncertainties}. In the context of machine learning, entropy is characterized as the disorder or uncertainty of the prediction model, or simply the level of `surprise' exhibited by the model when encountering particular data samples \cite{entropy}. The lower the value of entropy, the better. Equation \ref{enteq} is utilized in this study to calculate the entropy of our binary classification samples, where P is the prediction probability for the particular data under consideration.

\begin{equation}
Entropy =  \left(\emph{P} \times \log_2(\frac{1}{\emph{P}})\right) + \left((1-\emph{P})\times \log_2\frac{1}{(1-\emph{P})}\right)
\label{enteq}
\end{equation}

\vspace{0.5cm}

In our uncertainty analysis, we employed both variance and entropy metrics. Variance was computed across all four sequence categories, while entropy was determined for specific sequence encodings exhibiting higher variance compared to others. The top-performing GoogLeNet model was selected for prediction analysis. Images from four distinct categories were collected: PVP short sequences, PVP long sequences, non-PVP short sequences, and non-PVP long sequences. The average and variance of the prediction distribution were graphed and compared for each category. It was observed that the model exhibited higher confidence (lower prediction variance) when predicting non-PVP sequences compared to PVP sequences, and showed greater certainty for shorter sequences compared to longer ones. This uncertainty pattern was validated by employing various dropout rates (0.1, 0.2, 0.3) and randomly shuffling the sequences within each category. The figures presented utilize a dropout value of 0.2 for illustration purposes. Subsequently, Table \ref{mcd} lists the variance and corresponding MCD values for each category, indicating lower variance for shorter sequences and non-PVP predictions. Samples with the highest and lowest variance were selected from each category, and their mean prediction entropy was calculated, which is listed in Table \ref{meanentropy}. It was found that entropy values were lower for shorter sequences and non-PVP predictions, corroborating the observed uncertainty patterns.

\begin{table}[htbp]
\caption{Variance for PVP and non-PVP with different dropout rates}
\label{mcd}
\centering
\small
\begin{tabular*}{\columnwidth}{@{\extracolsep{\fill}}lccc@{}}
\toprule
\textbf{Categories} & \textbf{MCD 0.1} & \textbf{MCD 0.2} & \textbf{MCD 0.3} \\
\midrule
PVP (short)      & 0.06801 & 0.08558 & 0.05536 \\
non-PVP (short)  & 0.05914 & 0.07555 & 0.04860 \\
\midrule
PVP (long)       & 0.07276 & 0.09119 & 0.05926 \\
non-PVP (long)   & 0.05994 & 0.07663 & 0.04900 \\
\bottomrule
\end{tabular*}
\end{table}

\section{Discussion}
\label{sec:discussion}
The primary objective of this study was to investigate the viability of employing a highly efficient DNA sequence encoding technique for the more intricate task of phage protein classification and to assess whether the results would be equally effective as those obtained with DNA sequences. We leveraged a variety of state-of-the-art deep learning convolutional neural networks to classify the encoded images. Additionally, we conducted an uncertainty analysis on the most efficient model to evaluate its strengths and limitations.
Traditional machine learning methods for phage virion protein classification, such as Meta-iPVP \mbox{\cite{charoenkwan2020meta}} and PVPred-SCM\cite{charoenkwan2020pvpred}, rely on algorithms like random forest, genetic algorithms, and support vector machines. These approaches require manual feature curation, such as k-mer frequencies and probabilistic features, and generally yield lower accuracy compared to newer deep learning methods. 

\begin{table}[htbp]
\caption{Entropy for PVP and non-PVP samples with high and low variance}
\label{meanentropy}
\centering
\small
\begin{tabular*}{\columnwidth}{@{\extracolsep{\fill}}lcc@{}}
\toprule
\textbf{Categories} &
\textbf{Entropy} (\textbf{High Var}) &
\textbf{Entropy} (\textbf{Low Var}) \\
\midrule
PVP (short)      & 0.1122  & 0.1259 \\
non-PVP (short)  & 0.1369  & 0.1398 \\
\midrule
PVP (long)       & 0.2111  & 0.3943 \\
non-PVP (long)   & 0.06539 & 0.1716 \\
\bottomrule
\end{tabular*}
\end{table}
 
Deep learning techniques, where hidden features are learned by neural networks, have shown significant improvements in PVP classification. For instance, PhANN \cite{PhANN} demonstrates superior performance compared to earlier methods, but it still relies on inputting thousands of features into its predictive artificial neural network architecture, which can be computationally intensive and potentially less efficient.  Additionally, the study \cite{fang2022deepvp} mitigates the need for extensive feature extraction by using one-hot encoding and employing a more powerful 1-D CNN architecture for classification. While this method shows good accuracy, the results are inconsistent across other evaluation metrics.
Moreover, Shang et al. \cite{phaVIP} introduce the classification of FCGR-based encoded sequences using a transformer architecture, achieving high prediction performance on the dataset. However, this FCGR encoding tends to result in spatial feature loss \cite{akbari2022walkim}. 

\begin{table*}[t]
\centering
\caption{Qualitative comparison of existing methods and ProteoKnight}
\label{qual}
\small
\renewcommand{\arraystretch}{1.2}
\begin{tabular*}{\textwidth}{@{\extracolsep{\fill}}lllp{5cm}c@{}}
\toprule
\textbf{Work} & \textbf{Approach} & \textbf{Model} & \textbf{Feature} & \textbf{Image Data} \\
\midrule
PhANN & Deep learning & ANN & K-mer frequency & \ding{55} \\
DeePVP & Deep learning & 1-D CNN & One-hot encoding & \ding{55} \\
VirionFinder & Deep learning & CNN & One-hot encoding, biochemical properties & \ding{55} \\
Meta-iPVP & Machine learning & SVM, RF, NB, ANN & Probabilistic features & \ding{55} \\
PVPred-SCM & Statistical & SCM, Genetic Algorithm & Dipeptide composition & \ding{55} \\
PhaVIP & Deep learning & ViT & FCGR & \ding{51} \\
\textbf{ProteoKnight} & \textbf{Deep learning} & \textbf{CNN} & \textbf{Knight encoding} & \textbf{\ding{51}} \\
\bottomrule
\end{tabular*}
\end{table*}

Addressing these concerns, our approach, ProteoKnight, explores an image-based encoding method based on the DNA-walk algorithm, which potentially mitigates the issue of spatial loss \cite{akbari2022walkim} in the image-based representation of biological sequences. We chose to work with pre-trained CNN architectures, specifically GoogleNet, as they have demonstrated superior performance compared to pre-trained transformers for many classification tasks \cite{tay2021pre}. This, in turn, offers the advantage of leveraging existing knowledge, significantly reducing training time and computational needs, while also providing consistent results across various evaluation metrics by using a proven, high-efficiency DNA encoding method adapted for proteins. A qualitative comparison between various methods and our approach is highlighted in Table \ref{qual}. Our proposed technique for encoding and model selection yielded prediction performance comparable to the current state-of-the-art for binary classification. 


However, in the multi-class prediction task, our model exhibited room for further improvement, producing moderate outcomes with an accuracy range of 72\% to 78\%. This can be attributed to a characteristic of our suggested encoding approach, where dot points may converge at the same location when amino acids align, potentially leading to overlap, particularly when one dot is precisely positioned over another. While this characteristic did not significantly impact the less complex binary classification task, it posed a challenge for the more intricate multi-class classification scenario. Nevertheless, our encoding technique demonstrated promising potential, and with further refinements, it could potentially address the multi-class classification task more effectively.
The class-based uncertainty analysis revealed that our model demonstrated greater confidence when applied to non-PVP data compared to PVP data. This disparity may stem from the larger volume of training data available for the non-PVP category versus the PVP category, or it could be related to their distinctive underlying sequence compositions. These insights into model performance and uncertainty not only inform our understanding of the current approach but also guide our future research directions, setting the stage for more robust and versatile protein classification methodologies.

\section{Conclusion}
\label{sec:conclusion}
Bacteriophages play vital ecological roles and present promising therapeutic avenues against bacterial pathogens. Accurate classification of their structural proteins, i.e., phage virion proteins (PVPs), is therefore crucial across disciplines. This study introduces ProteoKnight, a novel image encoding methodology for PVP sequences. Overcoming spatial constraints of existing techniques like frequency chaos game representation, ProteoKnight enables high-accuracy binary PVP classification (GoogleNet - 89\% \& EfficientNet V2 - 91\%)  via pre-trained convolutional neural networks on a benchmark dataset. Furthermore, our uncertainty quantification through Monte Carlo dropout elucidates how prediction confidence varies across PVP classes and sequence lengths - pioneering insights for proteomic research. Collectively, ProteoKnight, coupled with efficient neural models, offers an accurate way to annotate important phage proteins, thereby setting the stage for exploiting the great medical and environmental benefits of bacteriophages through careful analysis of their viral proteins. 

While these results are promising, our research has also uncovered areas for further investigation and improvement. During the analysis, an atypical trend in prediction accuracy was observed for PVP sequences, with the model performing poorly when dropout was employed during inference. This phenomenon serves as an anchor for further investigation of uncertainty quantification for biological sequences. Regarding length-based uncertainty, the encoding limitation mentioned in the discussion section may elucidate the increased uncertainty observed in longer sequences compared to shorter ones, as longer sequence encodings are more susceptible to point overlaps. To address this issue, future research could focus on enhancing the knight encoding approach by extending it to a higher dimension through the segregation of frames for each amino acid or by optimizing the encoding's hyperparameters, such as radius, sphere/dot size, and so forth. This would enable the model to more effectively process the point representation of each amino acid residue, ensuring that no sequence information is compromised.

\section*{Credit authorship contribution statement}
\textbf{Md. Ishrak Khan:} Conceptualization, Software, Data curation, Formal analysis, Investigation, Validation, Visualization, Writing -- original draft, Writing -- review \& editing. 
\textbf{Abir Ahammed Bhuiyan:} Conceptualization, Software, Methodology, Investigation, Data curation, Writing -- original draft, Writing -- review \& editing. 
\textbf{Samiha Afaf Neha:} Visualization, Methodology, Investigation, Formal analysis, Validation, Writing -- original draft, Writing -- review \& editing. 
\textbf{Md. Khalilur Rahman:} Supervision, Project administration, Writing -- review \& editing. 
\textbf{Jannatun Noor:} Supervision, Project administration, Writing -- review \& editing.

\section*{Declaration of Competing Interest}
The authors declare that they have no known competing financial interests or personal relationships that could have appeared to influence the work reported in this paper.

\section*{Acknowledgment}
The authors would like to thank all individuals who provided valuable discussions and support during the course of this research.

\section*{Data and Code Availability}
This study utilizes the benchmark dataset established by Shang et al.\ \cite{phaVIP}.
The dataset was constructed by retrieving the latest sequence annotations up to
December 2022 from the RefSeq viral protein database, followed by necessary data
reconditioning procedures. The dataset is publicly available at
\url{https://phage.ee.cityu.edu.hk/download}.

The source code for ProteoKnight --- including the Knight Encoding scheme, dataset
construction, model training and evaluation, and the Monte Carlo Dropout uncertainty
analysis --- is publicly available on GitHub at
\url{https://github.com/eniac00/ProteoKnight}.

\bibliographystyle{elsarticle-num}
\bibliography{cas-refs}
\end{document}